\documentclass{article}

     \usepackage[preprint]{neurips_2021}

\usepackage[utf8]{inputenc} 
\usepackage[T1]{fontenc}    
\usepackage{hyperref}       
\usepackage{url}            
\usepackage{booktabs}       
\usepackage{amsfonts}       
\usepackage{nicefrac}       
\usepackage{microtype}      
\usepackage{xcolor}         

\usepackage{graphicx}

\usepackage{tikz}
\usepackage{float}
\usepackage{comment}
\usepackage{amsmath,amssymb} 
\usepackage{color, colortbl}
\usepackage{caption}
\usepackage{algorithm}
\usepackage{algpseudocode}
\usepackage{bbm}
\usepackage{subcaption}
\usepackage{csquotes}
\definecolor{LightCyan}{RGB}{255, 250, 232}

\usepackage[accsupp]{axessibility}  

\title{APP: Anytime Progressive Pruning}

\author{%
  Diganta Misra\thanks{Equal contribution} \\
  Mila - Quebec AI Institute, Landskape AI, UdeM\\
   \And
  Bharat Runwal\footnotemark[1] \\
  Landskape AI, IIT-Delhi \\
  \AND
  Tianlong Chen \\
  VITA, UT-Austin \\
  \And
  Zhangyang Wang \\
  VITA, UT-Austin \\
  \And
  Irina Rish \\
  Mila - Quebec AI Institute, UdeM\\
  \AND
  \texttt{\{diganta, bharat\}@landskape.ai}
}

\begin{document}

\maketitle

\begin{abstract}
With the latest advances in deep learning, there has been a lot of focus on the online learning paradigm due to its relevance in
practical settings. Although many methods have been investigated for optimal learning settings in scenarios where the data stream is continuous over time, sparse networks training in such settings have often been overlooked. In this paper, we explore the problem of training a neural network with a target sparsity in a particular case of online learning: the anytime learning at macroscale paradigm (ALMA). We propose a novel way of progressive pruning, referred to as \textit{Anytime Progressive Pruning} (APP); the proposed approach significantly outperforms the baseline dense and Anytime OSP models across multiple architectures and datasets under short, moderate, and long-sequence training. Our method, for example, shows an improvement in accuracy of $\approx 7\%$ and a reduction in the generalization gap by $\approx 22\%$, while being $\approx 1/3$ rd the size of the dense baseline model in few-shot restricted imagenet training. We further observe interesting nonmonotonic transitions in the generalization gap in the high number of megabatches-based ALMA. The code and experiment dashboards can be accessed at \url{https://github.com/landskape-ai/Progressive-Pruning} and \url{https://wandb.ai/landskape/APP}, respectively.
\end{abstract}

\section{Introduction}

\begin{figure}[t]
\centering
\includegraphics[width=0.99\textwidth]{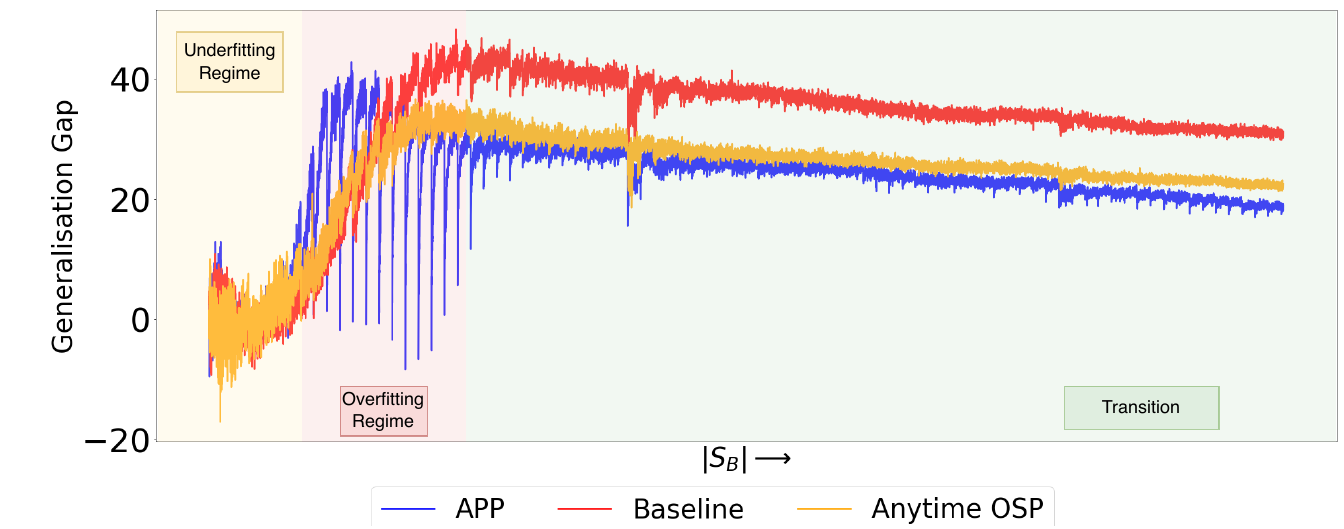}
\caption{\textbf{Non-Monotonic Transition:} Generalization gap as a function of the number of megabatches ($|S_B| = 100$) in the long-sequence ALMA setting with full replay for ResNet-50 backbone on the CIFAR-10 dataset. We observe that APP consistently demonstrates the \textit{lower} generalization gap compared to anytime OSP and baseline models. The curve for APP showcases high oscillation in the critical regime where the model starts overfitting, which can be attributed to the regularization effect induced by the pruning at the start of each megabatch.}
\label{fig:phase}
\end{figure}

Supervised learning has been one of the most well-studied learning frameworks for deep neural networks, where the learner is provided with a dataset $\mathcal{D}_{x,y}$ of samples($x$) and corresponding labels($y$); and the learner is expected to predict the label $y$ by learning on $x$ usually by estimating $p(y | x)$. In an offline learning environment \cite{ben1997online}, the learner has access to the complete dataset $\mathcal{D}_{x,y}$, while in a standard online learning setting \cite{sahoo2017online,bottou1998online} the data arrive in a stream over time, assuming that the rate at which samples arrive is the same as that of the learner's processing time to learn from them. There are several fine-grained types of learning from a stream of data, including, but not limited to, continuous learning \cite{van2019three,thrun1995lifelong,ring1998child}, active online learning \cite{baram2004online,settles2009active}, and anytime learning \cite{grefenstette1992approach,ramsey1994case}. In an anytime learning framework, the learner has to have good performance at any point in time, while gradually improving its performance over time upon observing new data that subsequently arrive. 

Anytime Learning at Macroscale(ALMA) \cite{caccia2021anytime} recently introduced a new subparadigm of learning inspired by anytime learning and transfer learning \cite{pan2009survey}. In ALMA, the time it takes for the model to be trained on a set of samples called a megabatch is significantly shorter than the interval between the arrival of two consecutive megabatch. Thus, ALMA studies the optimal waiting time that corresponds to the mega-batch size to ensure that the model is a good anytime learner. Caccia et al. \cite{caccia2021anytime} abstractly define a learner trained in an ALMA setting as:\newline
\blockquote{\emph{... a learner that i) produces high accuracy, ii) can make non-trivial predictions at any point in
time, while iii) limits its computational and memory resources}}

In this work, we are interested in exploring the training of sparse(pruned) neural networks in the ALMA setting. Pruning \cite{blalock2020state,luo2017thinet,wang2021emerging} of overparameterized deep neural networks has been studied for a long time. Pruning deep neural networks leads to a reduction in inference time and memory footprint. Pruning has gained prominence since the inception of the lottery ticket hypothesis (LTH) \cite{frankle2018lottery,frankle2019roy,frankle2019stabilizing,malach2020proving}, which demonstrated the existence of subnetworks (lottery tickets) within a dense network, which, when trained from random initialization in the same setting, will match or outperform the dense network. Although early pruning work focused exclusively on pruning weights after pretraining the dense model for a certain number of iterations, extensive research has recently been conducted on pruning the model at initialization, that is, finding the lottery ticket from a dense model at the start without pretraining the dense model \cite{lee2018snip,Wang2020PickingWT}. However, few studies \cite{chen2020long} have investigated the training of sparse(pruned) neural networks in online settings. Thus, our objective is to answer the following question: \newline

\textquote{\emph{Given a dense neural network and a target sparsity, what should be the optimal way of pruning the model in ALMA setting?}}\newline

In summary, our contributions can be summarized by the following four points.
\begin{enumerate}
    \item[$\ast$] We provide the first comprehensive study into pruning of deep neural networks in an ALMA setting. Specifically, we conclude through extensive empirical evaluation that progressive pruning consistently outperforms different baselines. We define the baselines used for comparison in Section \ref{sec: app}.
    \item[$\ast$] We therefore propose a novel approach of progressively pruning dense neural networks in the ALMA paradigm, which we term \textbf{Anytime Progressive Pruning}(APP). 
    \item[$\ast$] We further investigate the training dynamics of APP as compared to the baselines in ALMA setting with varied number of megabatches using CIFAR-10, CIFAR-100 and Restricted ImageNet datasets, and, observe non-monotonic transition graphs in their generalization gap during training. 
    \item[$\ast$] Furthermore, we do conclusive ablation studies to investigate the different types of pruners that are compatible with APP and one-shot pruning (OSP) models, along with studying the effect of replay. We conclude that APP outperforms OSP when all $t-1$ megabatches are replayed while training on the $t$-th megabatch; however, OSP models outperform APP models when no replay buffer is used.
\end{enumerate}

In the following section, we provide concrete insights into the motivation of the problem statement that we investigate in this paper, derived from the foundations of active learning and practical data acquisition(collection, annotation, and labeling).

\subsection{Motivation}

A well-accepted statement in deep learning states \textquote{\emph{Collection of unlabeled data is relatively easy, however, labeling is costly and difficult.}} This is arguably true because labeling or annotating data requires a human in the loop with extensive domain knowledge, which induces an additional cost in addition to the cost in the form of computing power required to train the learner. Active learning \cite{baram2004online,settles2009active,olsson2009literature,Liu_2021_ICCV,dimitrakakis2008cost} is a well-studied domain in machine learning that particularly investigates training of data-efficient models under cost constraints. Specifically, given a learner and a set of unlabeled data, an active learning algorithm will select particular samples to label via an oracle, under a predetermined cost budget, to maximize performance. This framework is not only limited to the labeling of unlabeled data, but can also be extended to label correction or reannotation of noisy labeled data. In \cite{bernhardt2021active}, the authors study optimal reannotation strategies under resource constraints to achieve a maximal performance gain, which they call active label cleaning. The authors of \cite{settles2008active} study the annotation times and costs of different data sets in the real world domain. They report the variation in cost and time required to label different sets of unlabeled data. 

Reiterating from the previous section, we are interested in understanding and finding the optimal strategy for training sparse neural networks given a target sparsity in the ALMA setting. Often, in industrial and practical scenarios, there exists a fixed initial period for the collection of data, which is subsequently labeled by an oracle. For this problem statement, we assume knowledge of the total number of samples that the learner will observe, which allows us to predetermine the required number of megabatches. We assume that the complete stream of data is already acquired but unlabeled and that the individual megabatches received in the stream by the learner are labeled over time by an oracle. This allows us to optimally select the wait time (megabatch size) and study the interesting properties of the training dynamics of models trained on these megabatches in an ALMA setting.

\section{Related Work}

\subsection{Pruning}
Pruning~\cite{lecun1990optimal,han2015deep} as one of the effective model compression techniques is widely explored in the field of efficient machine learning. It trims down the parameter redundancy in modern over-parameterized deep neural networks, aiming at substantial resource savings and unimpaired performance. Depending on the granularity of the removed network components, classical pruning methods can be categorized into unstructured~\cite{han2015deep,lecun1990optimal,han2015learning} and structural pruning~\cite{liu2017learning,zhou2016less}, where the former removes parameters irregularly and the latter discards substructures such as convolution filters or layers. In addition to the above post-training pruning, it can also be flexibly applied before network training, such as SNIP~\cite{lee2019snip}, GraSP~\cite{wang2020picking} and SynFlow~\cite{tanaka2020pruning} or during training~\cite{zhang2018adam,he2017channel}. The key factor in these methods is the estimation of the importance of pruning targets, which can be learned~\cite{zhang2018adam,he2017channel} using data-driven methods or approximated by some heuristics of the training dynamics, including weight magnitude~\cite{han2015deep}, gradient~\cite{molchanov2019importance}, hessian~\cite{lecun1990optimal}. 

Recent closely related work~\cite{chen2021long} defines pruning in sequential learning as a dynamical system and proposes two effective lifelong pruning algorithms to identify high-quality subnets, leading to superior trade-offs between efficiency and lifelong learning performance. Furthermore, \cite{golkar2019continual} prunes neurons with low activity and \cite{sokar2020spacenet} compresses the sparse connections of each task during training to overcome the problem of forgetting.

\subsection{Lifelong Learning}
Lifelong learning~\cite{ring1994continual,thrun1995lifelong,ring1998child,thrun1998lifelong} has gained increasing attention from the deep learning community. Numerous algorithms developed can be roughly divided into two categories: ($i$) one group of methods~\cite{wang2017growing,rosenfeld2018incremental,rusu2016progressive,aljundi2017expert,rebuffi2018efficient,mallya2018piggyback} accommodate newly added tasks/classes by accordingly growing the network capacity. However, it usually suffers from the explosive model size, which is proportional to the number of classes. ($ii$) the other group of approaches mainly takes advantage of advances in transfer learning~\cite{kemker2017fearnet,belouadah2018deesil}, where the quality of pre-trained embeddings plays an essential role. In particular, \cite{li2017learning,castro2018end,javed2018revisiting,rebuffi2017icarl,Belouadah_2019_ICCV,belouadah2020scail} adopt replay methods with some stored past training data to alleviate catastrophic forgetting in sequential learning. Furthermore, more follow-ups use imbalance learning techniques~\cite{he2009learning,buda2018systematic} or knowledge distillation regularizations~\cite{li2017learning,castro2018end,he2018exemplar,javed2018revisiting,rebuffi2017icarl,Belouadah_2019_ICCV,belouadah2020scail} to further improve its performance on all learned tasks.

\subsection{Revisiting ALMA}
\label{sec: alma}

In this section, we revisit the ALMA learning framework as conceptualized in \cite{caccia2021anytime}. Based on the reasoning provided in the original paper, we explicitly focus on classification problems. In ALMA, the model $f_\theta$ is provided with a stream of $S_B$ of $|S_B|$ consecutive batches of samples under the assumption that there exists an underlying data distribution $\mathcal{D}_{x,y}$ with input $x \in \mathbb{R}^{d}$ and target labels $y \in \{1,...,C\}$. Each megabatch $\mathcal{M}_{t}$ consists of $N \gg 0$ i.i.d. samples randomly drawn from $\mathcal{D}_{x,y}$, for $t \in \{1,...,S_B\}$. Therefore, the stream $S_B$ is the ordered sequence $S_B = \{\mathcal{M}_1,....,\mathcal{M}_{|S_B|}\}$ where $|S_B|$ represents the total number of megabatches in the stream. Thus, the model $f_\theta:\mathbb{R}^d \rightarrow \{1,...,C\}$ is trained by processing a \textit{mini-batch} of $n \ll N$ samples at a specified time of each mega-batch $\mathcal{M}_t$ and iterating multiple times over each mega-batch before having access to the next mega-batch. In ALMA, it is assumed that the rate at which megabatches arrive is slower than the training time of the model on each megabatch, and, therefore, the model can iterate over the megabatches at its disposal based on its discretion to maximize performance. ALMA can be considered a special case of continual learning (CL) or lifelong learning~\cite{ring1994continual,thrun1995lifelong,ring1998child,thrun1998lifelong}, whose data distribution across batches (or tasks) is considered stationary. Compared to CL, the difficulty in ALMA is fewer data in each learning stage, while the challenge in CL is the dynamic data distributions across different learning stages. Meanwhile, ALMA is also loosely relevant to online learning~\cite{saad1998online} with the key difference that ALMA receives large batch data sequentially rather than in a stream. 

In ALMA, one of the main aims was to study the effect of variation in waiting time, which directly corresponds to the size of each megabatch, i.e. how long one should wait to collect samples for a particular megabatch. Furthermore, the authors conducted a conclusive study using different baselines, two of them being (a) ensemble and (b) dynamic growing. In both cases, the complexity of the model parameters was gradually increased to allocate sufficient capacity to accommodate the newly arrived megabatches. However, in this paper, we investigate the effect of progressively decreasing the parametric complexity of the model through pruning and subsequently training a sparse neural network in an ALMA setting. 

\section{Anytime Progressive Pruning}
\label{sec: app}

\begin{figure}
\centering
\includegraphics[width=0.98\textwidth]{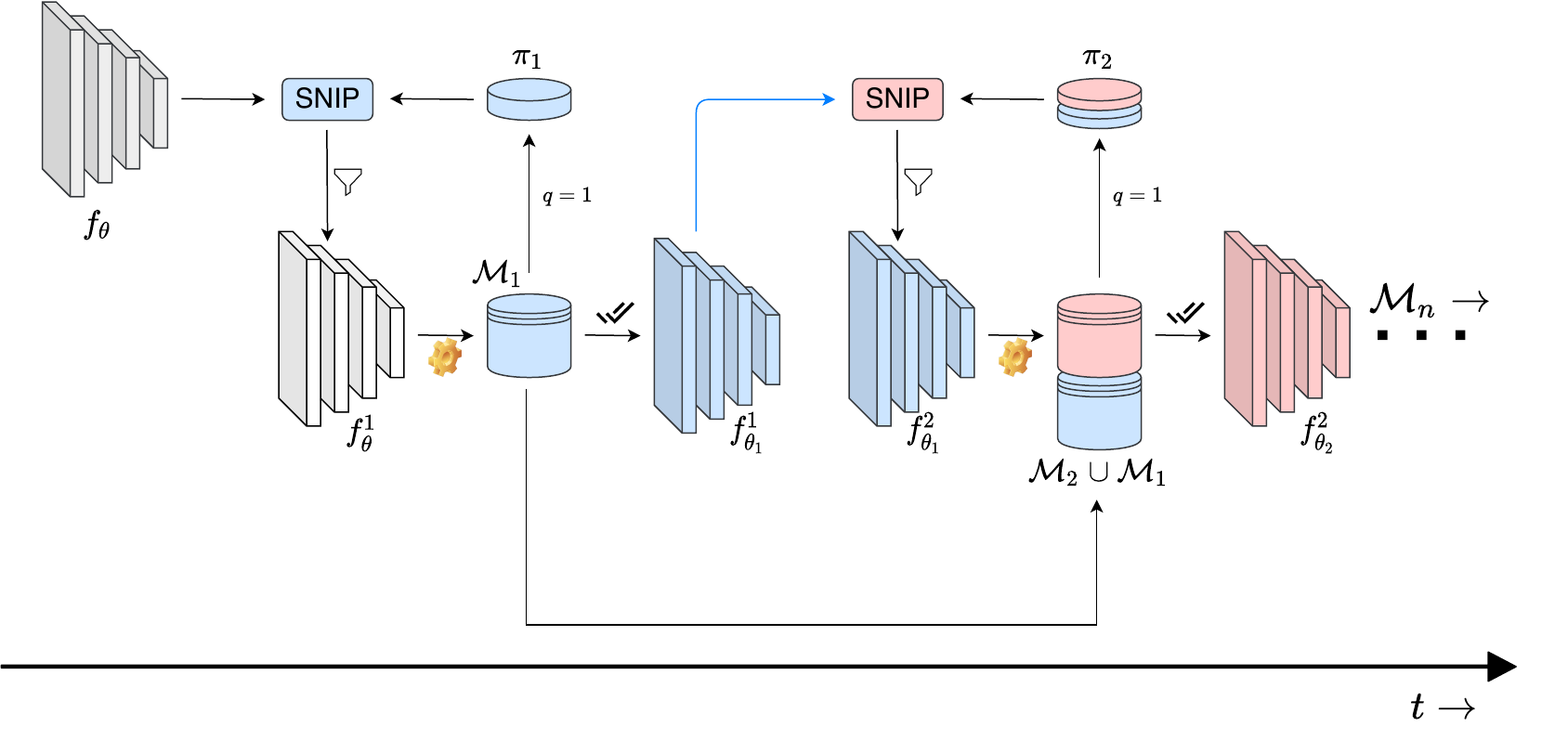}
\vspace{-4mm}
\caption{Overview of \textbf{Anytime Progressive Pruning} (APP) using full replay with a given randomly initialized dense model $f_\theta$ and $|S_B|$ total megabatches.}
\label{fig:app}
\end{figure}

In this section, we formally introduce our proposed method \textit{Anytime Progressive Pruning}(APP). As demonstrated in Figure \ref{fig:app}, given a randomly initialized dense neural network $f_\theta$, a target sparsity $0.8^\tau \times 100\%$, and the first megabatch $\mathcal{M}_{1} \in S_B$ containing $|\mathcal{M}_{1}|$ samples, we use $20\%$ random samples in $M_1$ denoted as $\pi_1$ and pass them to SNIP \cite{lee2018snip} together with $f_\theta$ and prune the model to $0.8^{\delta_1} \times 100\%$ in one iteration at initialization. After pruning, we take the pruned network $f_\theta^{1}$ and train it on $\mathcal{M}_1$ for $k$ epochs. For the next sequence $\mathcal{M}_2$, we first concatenate the entire previous megabatch $\mathcal{M}_1$ into the current megabatch $\mathcal{M}_2$ that gives $\mathcal{M}_1 \cup \mathcal{M}_2$ and then take the best performing checkpoint of the trained model on $\mathcal{M}_1$ - $f^1_{\theta_{1}}$ and again use $20\%$ random samples in $\mathcal{M}_1 \cup \mathcal{M}_2$ denoted as $\pi_2$ and pass them to SNIP to prune $f^1_{\theta_{1}}$ by further $0.8^{\delta_2} \times 100\%$ and use the resultant model $f^2_{\theta_{1}}$ to train it on $\mathcal{M}_1 \cup \mathcal{M}_2$. 

\begin{algorithm}
\caption{Training APP in the ALMA setting}\label{alg:app_algo}
\begin{algorithmic}[1]
\Require $f_\theta^{t=0}, \tau$, replay, $S_B \Longleftrightarrow \{\mathcal{M}_1,....,\mathcal{M}_{|S_B|}\}$
\State $t \gets 1$
\State $\delta \gets \{start=1,end=\tau,steps=|S_B|\}$ \Comment{Pruning states at each megabatches}
\While{$t \leq |S_B|$} \Comment{For each megabatch based training}
\State SNIP set($\pi_{t}$) $\gets \emptyset$ 
\If{replay}
    \State $\mathcal{M}_{t} \gets \bigcup\limits_{i=1}^{t} \mathcal{M}_{i}$
\Else{}
    \State $\mathcal{M}_{t} \gets \mathcal{M}_{t}$
\EndIf
\State pruning state $\gets {0.8^{\delta_t}}$ \Comment{Target Sparsity ($\times 100\%$) at each megabatch}
\State SNIP set $\gets \pi_{t} \subset \mathcal{M}_{t} \mid \frac{|\pi_{t}|}{|\mathcal{M}_t|} = 0.2$ 
\State $f_\theta^{t} \gets$ SNIP($f_\theta^{t-1}$, SNIP set, pruning state)
\State $f_\theta^t$\textit{.train(}$\mathcal{M}_t$\textit{)} \Comment{Fine-tune or retrain from scratch}
\EndWhile
\end{algorithmic}
\end{algorithm}

Thus, for each megabatch $\mathcal{M}_t \in S_B$, we construct the replay inclusive megabatch $\mathcal{M}_t$ by taking the union of all previous megabatches along with the current megabatch and then create a small sample set $\pi_t$ of size $0.2 \ast |\mathcal{M}_t|$ to be used to prune the model to $0.8^{\delta_t} \times 100\%$ sparsity. Here, $\delta_t$ is obtained from a predetermined list $\delta$ of uniformly spaced values that denote the target sparsity levels for each megabatch in the stream $S_B$. After pruning the model, we train it on the $\mathcal{M}_t$ megabatch and evaluate it on a holdout test set. 

\textbf{Note:}
\begin{enumerate}
    \item[*] The operator $|\cdot|$ denotes the size of a given set throughout this paper. 
    \item[*] $|\theta|$ denotes the number of trainable parameters in millions.
    \item[*] By default, APP always uses full replay buffer.
    \item[*] For all experiments, the scope of pruning was maintained to be \textit{Global}.
    \item[*] $0.8^\tau \times 100\%$ represents the fraction of weights left from the initial dense model post pruning and not the fraction of weights pruned, which would be denoted as $(1 - 0.8^\tau)\times 100\%$.
\end{enumerate}

To evaluate APP, we use primarily 2 baselines: 
\begin{enumerate}
    \item \textbf{Baseline}: This denotes the standard model (e.g., convolution neural network or transformer) at full parametric capacity trained and fine-tuned on all megabatches in the stream $S_B$ using stochastic gradient descent in an ALMA setting.
    \item \textbf{Anytime OSP}: This denotes one-shot pruning (OSP) to the target sparsity $0.8^\tau\ \times 100\%$ at the initialization of $f_\theta$ and then subsequently training on all mega-batches in the stream $S_B$ in an ALMA setting. Thus, anytime OSP models have the lowest parametric complexity since the start of training on the first megabatch in the stream $S_B$. We use the same pruner of choice (SNIP) by default for both APP and Anytime OSP. Similarly to APP, we prune the model at initialization using a small randomly selected subset $\pi_1$ of the first megabatch $\mathcal{M}_1$ of size $0.2 \ast |\mathcal{M}_1|$. 
\end{enumerate}

We use the following metric along with the test accuracy and the generalization gap to evaluate the methods specified above.
\begin{enumerate}
    \item \textbf{Cumulative Error Rate (CER)}: This can be defined by the following equation:
    \begin{equation}
        CER = \sum_{t = 1}^{S_B} \sum_{j = 1}^{|T_{x,y}|} \mathbbm{1}{(\mathcal{F}_t(x_j)\neq y_j)}
    \end{equation}
    Here, $T_{x,y}$ represents the held-out test set used for evaluation, $\mathcal{F}_t$ represents the trained model at $t$-th megabatch and $\mathcal{F}_t(x_j)$ represents the prediction on the $j$-th index sample of the test set $T_{x,y}$ compared to the true label for that sample $y_j$. CER provides strong insights into whether the learner is a good anytime learner, as it is expected to minimize CER at each megabatch training in the stream $S_B$.
\end{enumerate}

We follow the standard definition of the generalization gap as the difference between the training and the validation accuracy. This gives a notion of whether the model is over- or under-fitting. 

\section{Experiments}

In this section, we provide in-depth details on the experimental setup, the learning algorithms, and the data sets used in our empirical evaluation. We further discuss the training dynamics observed under the variation of $|S_B|$ and supplement our results with a visualization of the training curves. 

\subsubsection*{Datasets} \label{sec:data} We empirically evaluated APP, Anytime OSP, and Baseline models on three different data sets: (a) CIFAR-10 (C-10) \cite{krizhevsky2009learning} (b) CIFAR-100 (C-100) \cite{krizhevsky2009learning} and (c) Restricted Imagenet (balanced) \cite{robustness,tsipras2018robustness}. Both C-10 and C-100 consist of 50,000 training images and 10,000 test images, each of size 32 x 32, divided into 10 and 100 classes, respectively. Restricted ImageNet (balanced) is a subset of the original ImageNet data set \cite{russakovsky2015imagenet} consisting of 89517 training images and 3450 test images, each of 224 x 224 size divided into 14 classes consisting of five subclasses each. For our experimental analysis, we conduct benchmarks on both the 224 x 224 size version and additionally a 32 x 32 size version where we down-sample each image using bilinear interpolation. 

Taking into account the three datasets mentioned above, we construct benchmarks for the evaluation of APP as follows: (1) we randomly partition the data set into $|S_B|$ megabatches with an equal number of samples in each megabatch, (2) for each megabatch $\mathcal{M_T} \in S_B$, we partition it into a train set comprising 90\% samples in $\mathcal{M}_t$ and a validation set of the remaining 10\% samples in $\mathcal{M}_t$, (3) from each megabatch we randomly extract 20\% of the training data to build the set $|\pi_t|$ used for pruning via SNIP, and (4) we create the training pipeline where the learner is pruned using $\pi_t$ and subsequently trained on the megabatch at the current state for $k$ iterations. We keep a separate held-out test set, which is not seen by the learner during training, but is used to evaluate the model's performance after completion of training on each megabatch.

\subsubsection*{Models} \label{subsec: model} For our experiments, we use mainly four standard vision classifiers: (a) ResNet-18 \cite{he2016deep}, (b) ResNet-50 \cite{he2016deep}, (c) VGG-16 (with Batch Normalization) \cite{simonyan2014very}, and (d) Wide ResNet-50 \cite{zagoruyko2016wide}. We specifically picked these models because of their popularity in standard computer vision tasks and the extensiveness of the studies conducted on these models for various learning paradigms. However, for long-sequence-based ALMA (high number of megabatches) and restricted ImageNet experiments, we only use ResNet-50 as the model of choice. In addition, all models were trained from scratch, and no pre-training was used. 

\subsubsection*{Hyperparameters and learning setup}

Here, we describe in detail the experimental setups that were used for the reported results and discuss the difference in performance between APP, Anytime OSP, and baseline models in different scenarios. As mentioned above, for the experimental evaluation, we focus primarily on the task of image classification using the models defined in Subsection \ref{subsec: model}. For all experiments excluding a single VGG-16 + BN ablation study, we used a fixed target sparsity $\tau = 4.5$, which means, for all APP and Anytime OSP-based results, the model was pruned to have only 36.63\% remaining weights compared to the initial dense baseline network, which corresponds to $\approx \frac{1}{3}$rd model capacity post pruning. We hardcoded $\tau$ to 4.5, as we observe an inconsistency in performance for APP models at higher levels of sparsity, as reported in Table \ref{table:cifar_replay}. 

We use the following two learning setups for our empirical validation.
\begin{enumerate}
    \item \textbf{SGD with multi-step decay at $\mathcal{M}_1$ only}: All results reported in Table \ref{table:cifar_replay} were trained using Stochastic Gradient Descent (SGD) with a momentum of $0.9$ and an initial learning rate of 0.1, along with multistep decay of the learning rate by $\gamma = 0.1$ at the 91\textsuperscript{st} and 136\textsuperscript{th} epoch only for the first megabatch ($\mathcal{M}_1$). For all subsequent megabatches $\mathcal{M}_2 ... \mathcal{M}_t$, a constant learning rate of 0.001 was maintained. Each megabatch was trained for $182$ epochs except the $|S_B| = 25$ run reported in Table \ref{table:cifar_large}. 
    \item \textbf{SGD with cyclic multi-step decay at every $\mathcal{M}_t$}: All results in Table \ref{table:cifar_sgd_cyclic_norep} (excluding the results highlighted in \textcolor{yellow}{light yellow}), \ref{table:im32_few} and \ref{table:im32} were trained using SGD with the same initial parameters as described above. However, after completion of the training on each megabatch $\mathcal{M}_t$, the learning rate was reset to its initial state of $0.1$. 
\end{enumerate}

While we tested various pruning algorithms for the APP and Anytime OSP models such as SNIP \cite{lee2018snip}, magnitude pruning, random pruning, IMP \cite{frankle2018lottery}, and GraSP \cite{wang2020picking}, we use SNIP by default because we observe higher stability and training performance when coupled with APP. We use the default parameters for the pruning algorithms specified above and provide additional details in the supplementary section.

\subsection{Results}

\subsubsection{Ablation study with CIFAR-10 ALMA}

We conducted initial experiments to validate the design choice used in the default version of APP for all experiments conducted in this research. We used the following different versions of APP for the experiment:

\textbf{Note:} All experiments were carried out using a ResNet-50 backbone on CIFAR-10 using a fixed target sparsity of $\tau = 4.5$ with a total of 8 megabatches consisting of each 6250 samples. Furthermore, all reported results were obtained with a cyclic learning rate policy only for the first mega-batch $\mathcal{M}_1$, and subsequent mega-batches had a fixed learning rate.

\begin{enumerate}
    \item \textbf{APP Default:} This is the default version of APP adopted in all the experiments in this manuscript. The exact algorithm is defined in Section \ref{sec: app}.
    \item \textbf{APP + WD (1e-4):} This version of APP follows the same algorithm as that of the default, however, adds a weight decay of 1e-4 to the weight updates at each iteration.
    \item \textbf{APP Final:} In this version of APP, we apply the pruning at the end of each megabatch training, contrary to the default version where we prune the model at the beginning of each megabatch.
    \item \textbf{APP Warmup:} In this version of APP, we apply the pruning after a few warm-up epochs (20) at each megabatch, contrary to the default version where we prune the model at the beginning of each megabatch.
    \item \textbf{APP no replay SNIP:} In this version of APP, we construct the subset $\pi_t$ used by SNIP for pruning only from the current megabatch $\mathcal{M}_t$ and do not include any samples from the megabatch in the replay buffer $\mathcal{M}_1 .... \ \mathcal{M}_{t-1}$. 
\end{enumerate}

\begin{figure}[H]
\centering
\includegraphics[width=0.98\textwidth]{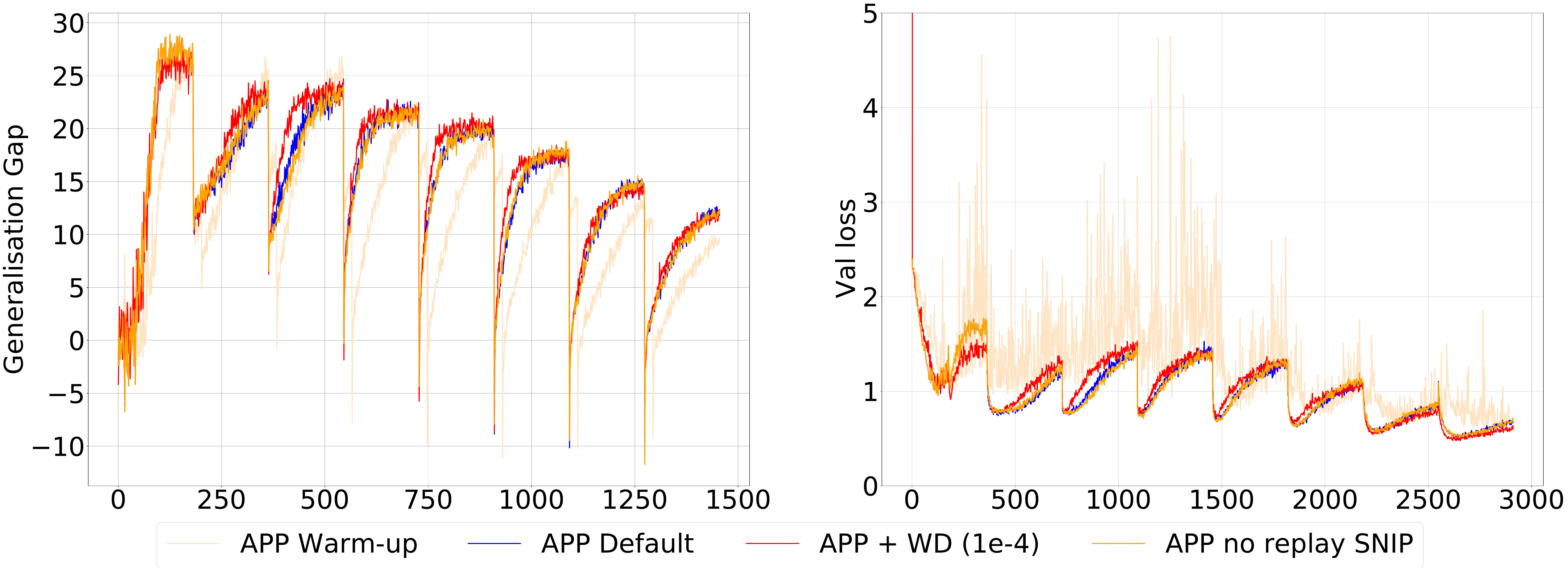}
\caption{From left to right: (a) Generalization gap for various versions of APP. (b) Validation loss for various versions of APP.}
\label{fig:ablation}
\end{figure}

As shown in Table \ref{table:Ablation}, APP + WD (1e-4) obtains the highest test accuracy, while APP warm-up has the lowest generalization gap compared to the default version of APP. However, we do not use weight decay by default due to inconsistent results across various models and settings. Although APP warm-up provides a drop of the generalization gap by a margin of $2.64\%$ compared to APP default, we do not use it as default due to the reduction in test accuracy by a margin of $1.8\%$ compared to APP default, and additionally the training of the same was extremely unstable, as shown in Fig. \ref{fig:ablation}. Furthermore, we note that the training collapses when used with APP Final version, while for the case of APP no-replay SNIP, we observe a drop in both the test accuracy and the generalization gap compared to APP default. 

\setlength{\tabcolsep}{4pt}
\begin{table}[H]
\begin{center}
\caption{Ablation study for the choice of APP design.}
\label{table:Ablation}
\resizebox{0.6\textwidth}{!}{\begin{tabular}{ccc}
\hline\noalign{\smallskip}
Method & Test Accuracy($\uparrow$) & Generalization Gap($\downarrow$)\\
\noalign{\smallskip}
\hline
\noalign{\smallskip}
APP Default & 84.65\% & 11.816\%\\
APP + WD (1e-4) & \textbf{85.6\%}(\textcolor{green}{+0.95 \%}) & 12.336\%(\textcolor{red}{+0.52 \%})\\
APP Final & 37.2\% & 65.333\%$^{\dagger}$ \\
APP Warm-up & 82.85\%(\textcolor{red}{-1.8 \%}) & \textbf{9.176\%}(\textcolor{green}{-2.64 \%}) \\
APP no replay SNIP & 83.35\%(\textcolor{red}{-1.3 \%}) & 11.976\%(\textcolor{red}{+0.16 \%}) \\
\hline
\end{tabular}}
\end{center}
\end{table}
\setlength{\tabcolsep}{1.4pt}

\subsubsection{Analysis of short sequence ALMA ($|S_B| = 8$)}

We start by analyzing the results reported in Table \ref{table:cifar_replay}. All experiments were carried out using full replay ($\mathcal{M}_{t} \gets \bigcup\limits_{i=1}^{t} \mathcal{M}_{i}$) for a total of 8 megabatches ($|S_B| = 8$) with each megabatch containing $|\mathcal{M}_t| = 6250$ samples. 

For ResNet-18 trained on C-10, we observe that APP (SNIP) decreases the test accuracy by $0.68\%$ and decreases the generalization gap by $6.718\%$ compared to the baseline model. Although the Anytime OSP (SNIP) model outperforms APP (SNIP) by a small margin of $0.21\%$, the former has a significantly higher generalization gap compared to the latter by $6.867\%$. For C-100, we observed a strong improvement in APP (SNIP) in test accuracy compared to the baseline by a margin of $3.85\%$, while the generalization gap was drastically reduced by $30.187\%$. Similar improvements were observed for the CER metric as reported in the table and the improvements were consistent when using magnitude-based pruning for APP. 

For ResNet-50, we observed an even greater performance improvement for APP (SNIP) compared to the baseline and Anytime OSP models in all metrics: test accuracy, CER, and generalization gap. For example, in C-10, the use of APP (SNIP) improved the test accuracy by $5.46\%$, decreased the CER by $1278$, and decreased the generalization gap by $7.393\%$ while the use of Anytime OSP (SNIP) resulted in a decrease in the test accuracy by $3.1\%$, an increase in the CER by $4268$, and an increase in the generalization gap by $3.11\%$ compared to the baseline model. We observe for C-10 that the use of APP (SNIP) with a small weight decay of $1e-4$ results in an improvement in the test accuracy by $0.95\%$ compared to the version without weight decay. However, we did not conduct extensive studies to validate the improvement caused by weight decay, as it is beyond the scope of our experimental evaluation. We observe a similar improvement in performance from baseline when coupled with weight decay.

The improvement in performance is also observed in Wide ResNet-50, where for C-10, APP (SNIP) outperforms the baseline model in test accuracy by $11.04\%$, reduces CER by $6095$ and reduces the generalization gap by $12.007\%$. However, for VGG-16 with batch normalization, we did not observe a significant improvement in performance over the baseline model compared to its Anytime OSP counterpart. 

For all experiments, we observed strong results for APP when used with SNIP and magnitude-based pruning. In our observations, while Anytime OSP is stable and compatible with other pruning methods such as random pruning and GraSP, APP causes a significant loss in performance when used with the same. This is the reason why we chose to fix SNIP as the pruner of choice for APP. We observe that APP with random pruning and GraSP continues to perform at par with its SNIP and magnitude-based pruning counterparts for the initial megabatches, but with increasing sparsity, causes a detrimental effect on the accuracy curves, as shown in Fig. \ref{fig:train(a)}. Furthermore, for VGG-16 with batch normalization, we conducted an experiment to study the effect of high sparsity for the training of C-10 where we set $\tau = 13$, which implies that the model had $\approx 5.5\%$ remaining parameters after pruning. However, we observed that APP causes a significant reduction in performance at this high level of sparsity.

\setlength{\tabcolsep}{4pt}
\begin{table}[H]
\begin{center}
\caption{Results on ALMA of C-10 and C-100. $\dagger$ denotes unstable training. WD denotes a weight decay of 1e-4 being used during training.}
\label{table:cifar_replay}
\resizebox{\textwidth}{!}{\begin{tabular}{cccccccccc}
\hline\noalign{\smallskip}
Backbone & Method & Pruner & $|\theta|$($\downarrow$) & \multicolumn{2}{c}{Test Accuracy($\uparrow$)} &
\multicolumn{2}{c}{CER($\downarrow$)} &
\multicolumn{2}{c}{Generalization Gap($\downarrow$)}\\
& & & & CIFAR-10 & CIFAR-100 & CIFAR-10 & CIFAR-100 & CIFAR-10 & CIFAR-100\\
\noalign{\smallskip}
\hline
\noalign{\smallskip}
ResNet-18 & Baseline & - & 11.51 M & 86.37\% & 54.44\%
& 14618 & 42535
& 13.64\% & 47.08\%\\
- & Baseline (WD) & - & 11.51 M & \textbf{88.75\%}(\textcolor{green}{+2.38 \%}) & 55.2(\textcolor{green}{+0.76 \%}) & 11840(\textcolor{green}{-2778}) & 42269 (\textcolor{green}{-266})& 11.0\%(\textcolor{green}{-2.64 \%}) & 46.553\%(\textcolor{green}{-0.527 \%}) \\
- & Anytime OSP & SNIP \cite{lee2018snip} & 4.09 M & \textbf{85.9\%}(\textcolor{red}{-0.47 \%}) & 54.09\%(\textcolor{red}{-0.35 \%})&
14276 (\textcolor{green}{-342}) & 42785 (\textcolor{red}{+250}) &
13.789\%(\textcolor{red}{+0.149 \%}) & 47.753\%(\textcolor{red}{+0.673 \%})\\
- & APP & SNIP \cite{lee2018snip} & 4.09 M & 85.69\%(\textcolor{red}{-0.68 \%}) & \textbf{58.29\%}(\textcolor{green}{+3.85 \%}) 
& \textbf{13476} (\textcolor{green}{-1142}) & \textbf{42442} (\textcolor{green}{-93})
&\textbf{6.922\%}(\textcolor{green}{-6.718 \%}) & \textbf{16.893\%}(\textcolor{green}{-30.187 \%})\\
- & Anytime OSP & Magnitude & 4.09 M & \textbf{86.2\%}(\textcolor{red}{-0.17 \%}) & 54.06(\textcolor{red}{-0.38 \%}) & 
\textbf{14486} (\textcolor{green}{132}) & 42090 (\textcolor{green}{-445}) &
13.611\%(\textcolor{green}{-0.029 \%}) & 47.94\%(\textcolor{red}{+0.86 \%})\\
- & APP & Magnitude & 4.09 M & 85.58\%(\textcolor{red}{-0.79 \%}) & \textbf{58.07\%}(\textcolor{green}{+3.63 \%})
& 16109 (\textcolor{red}{+1491}) & \textbf{41966} (\textcolor{green}{-569})
&\textbf{10.76\%}(\textcolor{green}{-2.88 \%}) & \textbf{22.676\%}(\textcolor{green}{-24.404 \%})\\
\hline
ResNet-50 & Baseline & - & 23.5 M & 79.19\% & 44.6\% & 
19221 & \textbf{49241} &
19.209\%  & 56.631\%  \\
- & Baseline (WD) & - & 23.5 M & \textbf{83.15 \%}(\textcolor{green}{+3.96 \%}) & 44.66\%(\textcolor{green}{+0.06 \%}) & 18143 (\textcolor{green}{-1078}) & 49879 (\textcolor{red}{+638}) & 16.991\%(\textcolor{green}{-2.218 \%}) & 57.233\%(\textcolor{red}{+0.602 \%}) \\
- & Anytime OSP & SNIP \cite{lee2018snip} & 8.6 M & 76.09\%(\textcolor{red}{-3.1 \%}) & 41.07\%(\textcolor{red}{-3.53 \%}) &
23489 (\textcolor{red}{+4268}) & 51829 (\textcolor{red}{+2588}) &
22.32\%(\textcolor{red}{+3.11 \%}) & 59.853\%(\textcolor{red}{+3.222 \%})\\
- & APP & SNIP \cite{lee2018snip} & 8.6 M & \textbf{84.65\%}(\textcolor{green}{+5.46 \%}) & \textbf{52.01\%}(\textcolor{green}{+7.41 \%})
& 17943 (\textcolor{green}{-1278}) & \textbf{48164} (\textcolor{green}{-1077}) &
\textbf{11.816\%}(\textcolor{green}{-7.393 \%}) & \textbf{23.002\%}(\textcolor{green}{-33.629 \%})\\
- & APP (WD) & SNIP \cite{lee2018snip} & 8.6 M & \textbf{85.6\%}(\textcolor{green}{+6.41 \%}) & - & & - & 12.336\%(\textcolor{green}{-6.873 \%})  & - \\
- & Anytime OSP & Magnitude & 8.6 M & 78\%(\textcolor{red}{-1.19 \%}) & 45.06\%(\textcolor{green}{+0.46 \%}) &
 21365 (\textcolor{red}{+2144}) & 48859 (\textcolor{green}{-382}) &
19.83\%(\textcolor{red}{+0.621 \%}) & 56.356\%(\textcolor{green}{-0.275 \%})\\
- & APP & Magnitude & 8.6 M & \textbf{83.63\%}(\textcolor{green}{+4.44 \%}) & \textbf{51.94\%}(\textcolor{green}{+7.34 \%}) 
& 19078 (\textcolor{green}{-143}) & \textbf{48032} (\textcolor{green}{-1209})
&\textbf{6.913\%}(\textcolor{green}{-12.296 \%}) & \textbf{22.871\%}(\textcolor{green}{-33.76 \%})\\
- & Anytime OSP & IMP & 8.6 M & 78.99\%(\textcolor{red}{-0.2 \%}) & - 
&- & -
& 20.227\%(\textcolor{red}{+1.018 \%}) & -\\
- & APP & IMP & 8.6 M & \textbf{83.63\%}(\textcolor{green}{+4.44 \%}) & - 
&- & -
&\textbf{6.913\%}(\textcolor{green}{-12.296 \%}) & -\\
- & Anytime OSP & Random & 8.6 M & \textbf{75.59\%}(\textcolor{red}{-3.6 \%}) & \textbf{47.07\%}(\textcolor{green}{+2.47 \%}) 
& 23745 (\textcolor{red}{+4524}) & \textbf{47646} (\textcolor{green}{-1595})
& 23.333\%(\textcolor{red}{+4.124 \%}) & \textbf{54.787\%}(\textcolor{green}{-1.844 \%})\\
- & APP & Random & 8.6 M & 62.63\%(\textcolor{red}{-16.56 \%}) & ${1.67\%}^{\dagger}$ 
& 24390 (\textcolor{red}{+5169}) & 55163 (\textcolor{red}{+5922})
& ${7.373\%}^{\dagger}$  & ${35.004\%}^{\dagger}$\\
- & Anytime OSP & GraSP \cite{wang2020picking}& 8.6 M & \textbf{83.26\%}(\textcolor{green}{+4.07 \%}) & \textbf{46.06\%}(\textcolor{green}{+1.46 \%})  
& 17442 (\textcolor{green}{-1779}) & \textbf{49244} (\textcolor{red}{+3})
&\textbf{16.862\%}(\textcolor{green}{-2.347 \%})  & \textbf{55.52\%}(\textcolor{green}{-1.111 \%})\\
- & APP & GraSP \cite{wang2020picking}& 8.6 M & $10.0\%^{\dagger}$ & $2.04\%^{\dagger}$ 
& 33415 (\textcolor{red}{+14194}) & 56254 (\textcolor{red}{+7013})
& ${0.1556\%}^{\dagger}$  & ${35.582\%}^{\dagger}$\\
\hline
Wide ResNet-50-2 & Baseline & - & 68.9 M & 74.45\%  & 47.42\% &
25299 & \textbf{47273} &
24.796\%  & 53.996\%\\
- & Baseline (WD) & - & 68.9 M & 84.28\%(\textcolor{green}{+9.83 \%})  & 51.31\%(\textcolor{green}{+3.89 \%}) & 17782 (\textcolor{green}{-7517}) & 45232 (\textcolor{green}{-2041}) & 14.578\%(\textcolor{green}{-10.218 \%}) & 50.718\%(\textcolor{green}{-3.278 \%}) \\
- & Anytime OSP & SNIP \cite{lee2018snip} & 25.2 M & 79.33\%(\textcolor{green}{+4.88 \%}) & \textbf{49.22\%}(\textcolor{green}{+1.8 \%}) & 
19815 (\textcolor{green}{-5484}) & \textbf{46052} (\textcolor{green}{-1221}) &
19.724\%(\textcolor{green}{-5.072 \%})  & 53.096\%(\textcolor{green}{-0.9 \%})\\
- & APP & SNIP \cite{lee2018snip} & 25.2 M & \textbf{85.49\%}(\textcolor{green}{+11.04 \%}) & 48.18\%(\textcolor{green}{+0.76 \%}) &
\textbf{19204} (\textcolor{green}{-6095}) & 48579 (\textcolor{red}{+1306}) &
\textbf{12.789\%}(\textcolor{green}{-12.007 \%}) & \textbf{38.64\%}(\textcolor{green}{-15.356 \%})\\
- & Anytime OSP & Magnitude & 25.2 M & 76.05\%(\textcolor{green}{+1.6 \%}) & 49.48\%(\textcolor{green}{+2.06 \%}) &
23900 (\textcolor{green}{-1399}) & \textbf{46174} (\textcolor{green}{-1099}) &
22.409\%(\textcolor{green}{-2.387 \%})  & 52.538\%(\textcolor{green}{-1.458 \%})\\
- & APP & Magnitude & 25.2 M & \textbf{85.28\%}(\textcolor{green}{+10.83 \%}) & \textbf{54.42\%}(\textcolor{green}{+7.0 \%})  &
\textbf{18675} (\textcolor{green}{-6624}) & 46697 (\textcolor{green}{-576})&
\textbf{12.291\%}(\textcolor{green}{-12.545 \%}) & \textbf{43.096\%}(\textcolor{green}{-10.9 \%})\\
- & Anytime OSP & Random & 25.2 M & \textbf{81.43\%}(\textcolor{green}{+6.98 \%}) & \textbf{44.5\%}(\textcolor{red}{-2.92 \%}) &
\textbf{18567} (\textcolor{green}{-6732}) & 49195 (\textcolor{red}{+1922}) &
\textbf{18.396\%}(\textcolor{green}{-6.4 \%}) & \textbf{57.393\%}(\textcolor{red}{+3.397 \%})\\
- & APP & Random & 25.2 M & 53.94\%(\textcolor{red}{-20.51 \%}) & 39.45\%(\textcolor{red}{-7.97 \%}) &
25673 (\textcolor{green}{-374}) & \textbf{48929} (\textcolor{red}{+1656}) &
${67.84\%}^{\dagger}$  & ${17.324\%}^{\dagger}$\\
- & Anytime OSP & GraSP \cite{wang2020picking}& 25.2 M & \textbf{81.49\%}(\textcolor{green}{+7.04 \%})  & \textbf{48.54\%}(\textcolor{green}{+1.12 \%})  &
\textbf{18452} (\textcolor{green}{-6847}) & - &
\textbf{17.311\%}(\textcolor{green}{-7.485 \%})& 53.902\%(\textcolor{red}{-0.094 \%}) \\
- & APP & GraSP \cite{wang2020picking}& 25.2 M & ${10.78\%}^{\dagger}$ & 26.27(\textcolor{red}{-21.15 \%}) &
29621 (\textcolor{red}{+4322}) & 55129 (\textcolor{red}{+7856}) &
${63.273\%}^{\dagger}$ & ${41.589\%}^{\dagger}$ \\
\hline
VGG-16-BN & Baseline & - & 138.42 M & 87.57\% & 53.52\%  & 12412 & 42410
& 11.747\%  & 48.329\%\\
- & Baseline (WD) & - & 138.42 M & \textbf{88.29\%}(\textcolor{green}{+0.72 \%}) & 54.85\%(\textcolor{green}{+1.33 \%}) & \textbf{11828} (\textcolor{green}{-584}) & 41122 (\textcolor{green}{-1288}) & 11.451\%(\textcolor{green}{-0.296 \%}) & 45.767\%(\textcolor{green}{-2.568 \%}) \\
- & Anytime OSP & SNIP \cite{lee2018snip} & 50.6 M & \textbf{87.59\%}(\textcolor{green}{+0.02 \%}) & 52.51\%(\textcolor{red}{-1.01 \%}) &
\textbf{12374} (\textcolor{green}{-38}) & 42575 (\textcolor{red}{+165}) &
12.24\%(\textcolor{red}{+0.493 \%})  & 47.811\%(\textcolor{green}{-0.518 \%})\\
- & APP & SNIP \cite{lee2018snip} & 50.6 M & 86.76\%(\textcolor{red}{-0.81 \%}) & \textbf{55.31\%}(\textcolor{green}{+1.79 \%}) &
12782 (\textcolor{red}{+370}) & \textbf{41285} (\textcolor{green}{-1125}) &
\textbf{10.113\%}(\textcolor{green}{-1.634 \%}) & \textbf{30.942\%}(\textcolor{green}{-17.387 \%})\\
- & Anytime OSP & SNIP \cite{lee2018snip} & 7.61 M & \textbf{86.75\%}(\textcolor{red}{-0.82 \%}) & - &
\textbf{13141} (\textcolor{red}{+729}) & - &
11.067\%(\textcolor{green}{-0.68 \%})  & - \\
- & APP & SNIP \cite{lee2018snip} & 7.61 M & 59.5\%(\textcolor{red}{-28.07 \%}) & - & 
20073 (\textcolor{red}{+7661})& - &
$-0.9733\%^{\dagger}$ & - \\
- & Anytime OSP & Magnitude & 50.6 M & \textbf{87.33\%}(\textcolor{red}{-0.24 \%}) &  53.27\%(\textcolor{red}{-0.25 \%}) & 
\textbf{12551} (\textcolor{red}{+139}) & \textbf{42306} (\textcolor{green}{-104}) &
12.476\%(\textcolor{red}{+0.729 \%})  & 47.996\%(\textcolor{green}{-0.333 \%})\\
- & APP & Magnitude & 50.6 M & 86.04\%(\textcolor{red}{-1.57 \%}) & \textbf{54.59\%}(\textcolor{green}{+1.07 \%}) &
12943 (\textcolor{red}{+531}) & 42310 (\textcolor{green}{-100}) &
9.862\%(\textcolor{green}{-1.885 \%}) & \textbf{22.369\%}(\textcolor{green}{-25.96 \%})\\
- & Anytime OSP & Random & 50.6 M & \textbf{87.49\%}(\textcolor{red}{-0.08 \%}) & \textbf{53.82\%}(\textcolor{green}{+0.3 \%}) &
\textbf{12539} (\textcolor{red}{+127}) & \textbf{41739} (\textcolor{green}{-671}) &
\textbf{12.533\%}(\textcolor{red}{+0.786 \%})  & \textbf{46.669\%}(\textcolor{green}{-1.66 \%}) \\
- & APP & Random & 50.6 M & 68.56\%(\textcolor{red}{-19.01 \%}) & 35.01\%(\textcolor{red}{-18.51 \%}) & 
16760 (\textcolor{red}{+4348}) & 46427 (\textcolor{red}{+4017}) &
${3.258\%}^{\dagger}$  & ${42.362\%}^{\dagger}$\\
- & Anytime OSP & GraSP \cite{wang2020picking} & 50.6 M & \textbf{87.04\%}(\textcolor{red}{-0.53 \%}) & \textbf{54.55\%}(\textcolor{green}{+1.03 \%}) &
\textbf{12945} (\textcolor{red}{+533}) & \textbf{41449} (\textcolor{green}{-961}) &
13.476\%(\textcolor{red}{+1.729 \%})  & \textbf{47.189\%}(\textcolor{green}{-1.14 \%})\\
- & APP & GraSP \cite{wang2020picking} & 50.6 M & ${14.57\%}^{\dagger}$ & ${26.27\%}^{\dagger}$ &
24131 (\textcolor{red}{+11719}) & 48888 (\textcolor{red}{+6478}) &
${56.624\%}^{\dagger}$  & ${41.589\%}^{\dagger}$\\
\hline
\end{tabular}}
\end{center}
\end{table}
\setlength{\tabcolsep}{1.4pt}

Compared to the results reported in Table \ref{table:cifar_replay}, we observe in Table \ref{table:cifar_sgd_cyclic} that using the cyclic learning rate policy at each megabatch $\mathcal{M}_t$ significantly improves performance for the three models, the baseline, Anytime OSP, and APP. For example, for ResNet-50, we note an improvement in test accuracy for the baseline model by a margin of $6.51\%$ compared to the baseline model trained with the cyclic learning rate policy only for the first megabatch $\mathcal{M}_1$ as reported in Table \ref{table:cifar_replay}. APP consistently outperforms the baseline and anytime OSP models for each experiment conducted on CIFAR-100 with a significant drop in the generalization gap observed for the four backbones used. 

\setlength{\tabcolsep}{4pt}
\begin{table}[t]
\begin{center}
\caption{Results on anytime learning of CIFAR-10 and CIFAR-100 using 8 mega-batches with 6250 samples per mega-batch with replay. All APP and Anytime OSP experiments used Snip \cite{lee2018snip} as the pruner of choice. All experiments also used an SGD + cyclic multistep decay LR at every $\mathcal{M}_t$.}
\label{table:cifar_sgd_cyclic}
\resizebox{\textwidth}{!}{\begin{tabular}{ccccccccc}
\hline\noalign{\smallskip}
Backbone & Method & $|\theta|$($\downarrow$) & \multicolumn{2}{c}{Test Accuracy($\uparrow$)} &
\multicolumn{2}{c}{CER($\downarrow$)} &
\multicolumn{2}{c}{Generalization Gap($\downarrow$)}\\
& & & CIFAR-10 & CIFAR-100 & CIFAR-10 & CIFAR-100 & CIFAR-10 & CIFAR-100\\
\noalign{\smallskip}
\hline
\noalign{\smallskip}
ResNet-18 & Baseline & 11.51 M & \textbf{91.43\%} & 60.39\% & 10545 & 38771 & \textbf{8.098\%} & 41.033\%\\
- & Anytime OSP & 4.09 M & 90.56\%(\textcolor{red}{-0.87 \%}) & 60.44\%(\textcolor{green}{+0.05 \%}) & 11255 (\textcolor{red}{+710}) & 38755 (\textcolor{green}{-16}) & 8.778\%(\textcolor{red}{+0.68 \%}) & 40.567\%(\textcolor{green}{-0.472 \%}) \\
- & APP & 4.09 M & 90.06\%(\textcolor{red}{-1.37 \%}) & \textbf{63.61\%}(\textcolor{green}{+3.22 \%}) & \textbf{10419} (\textcolor{green}{-126}) & \textbf{37048} (\textcolor{green}{-1723}) & 8.442\%(\textcolor{red}{+0.344 \%}) & \textbf{26.922\%}(\textcolor{green}{-14.111 \%})\\
\hline
ResNet-50 & Baseline & 23.5 M & 85.7\% & 46.91\% & 16821 & 49486 & 14.289\% & 54.878\% \\
 - & Anytime OSP & 8.6 M & 88.61\%(\textcolor{green}{+2.91 \%}) & 53.76\%(\textcolor{green}{+6.85 \%}) & 14209 (\textcolor{green}{-2612}) & 45092 (\textcolor{green}{-4394}) & 11.336\%(\textcolor{green}{-2.953 \%}) & 49.182\%(\textcolor{green}{-5.696 \%})\\ 
 - & APP & 8.6 M & \textbf{90.89\%}(\textcolor{green}{+5.19 \%}) & \textbf{64.88\%}(\textcolor{green}{+17.97 \%}) & \textbf{12294} (\textcolor{green}{-4527}) & \textbf{39559} (\textcolor{green}{-9927}) & \textbf{9.24\%}(\textcolor{green}{-5.049 \%}) & \textbf{34.387\%}(\textcolor{green}{-20.491 \%}) \\
\hline
Wide ResNet-50-2 & Baseline & 68.9 M & 89.65\% & 52.73\% & 13471 & 44866 & 10.398\% & 49.931\%\\
- & Anytime OSP & 25.2 M & 87.5\%(\textcolor{red}{-2.15 \%}) & 48.05\%(\textcolor{red}{-4.68 \%}) & 15717 (\textcolor{red}{+2246}) & 48978 (\textcolor{red}{+4112}) & 12.109\% (\textcolor{red}{+1.711 \%}) & 53.64\%(\textcolor{red}{+3.709 \%}) \\
- & APP & 25.2 M & \textbf{92.02\%}(\textcolor{green}{+2.37 \%}) & \textbf{66.24\%}(\textcolor{green}{+13.51 \%}) & \textbf{12808} (\textcolor{green}{-663}) & \textbf{40327} (\textcolor{green}{-4539}) & \textbf{7.976\%}(\textcolor{green}{-2.422 \%}) & \textbf{34.791\%}(\textcolor{green}{-15.14 \%}) \\
\hline
VGG-16-BN & Baseline & 138.42 M & \textbf{91.53\%} & 59.06\% & \textbf{9950} & 38615 & 9.318\% & 42.667\% \\
- & Anytime OSP & 50.6 M & 90.63\%(\textcolor{red}{-0.9 \%}) & 57.98\%(\textcolor{red}{-1.08 \%}) & 10236 (\textcolor{red}{+286}) & 39024 (\textcolor{red}{+409}) & 9.187\%(\textcolor{green}{-0.131 \%}) & 43.102\%(\textcolor{red}{+0.435 \%})\\
- & APP & 50.6 M & 89.82\%(\textcolor{red}{-1.71 \%}) & \textbf{62.51\%}(\textcolor{green}{+3.45 \%}) & 10171 (\textcolor{red}{+221}) & \textbf{36831} (\textcolor{green}{-1784})& \textbf{8.967\%}(\textcolor{green}{-0.351 \%}) & \textbf{33.293\%}(\textcolor{green}{-9.374 \%}) \\
\hline
\end{tabular}}
\end{center}
\end{table}
\setlength{\tabcolsep}{1.4pt}

\setlength{\tabcolsep}{4pt}
\begin{table}[H]
\begin{center}
\caption{Results on ALMA of C-10 and C-100 without replay using SGD with cyclic multistep decay at every $\mathcal{M}_t$. \textcolor{yellow}{Rows} highlighted in \textcolor{yellow}{light yellow} represent runs done with SGD with multistep decay at $\mathcal{M}_1$ only.}
\label{table:cifar_sgd_cyclic_norep}
\resizebox{\textwidth}{!}{\begin{tabular}{cccccccccc}
\hline\noalign{\smallskip}
Backbone & Method & Pruner & $|\theta|$($\downarrow$) & \multicolumn{2}{c}{Test Accuracy($\uparrow$)} &
\multicolumn{2}{c}{CER($\downarrow$)} &
\multicolumn{2}{c}{Generalization Gap($\downarrow$)}\\
& & & & CIFAR-10 & CIFAR-100 & CIFAR-10 & CIFAR-100 & CIFAR-10 & CIFAR-100\\
\noalign{\smallskip}
\hline
\noalign{\smallskip}
ResNet-18 & Baseline & - & 11.51 M & 87.42\% & 53.4\% & \textbf{12834} & \textbf{41919} & 13.76\% & \textbf{45.44\%} \\
- & Anytime OSP & SNIP \cite{lee2018snip}& 4.09 M & \textbf{87.72\%}(\textcolor{green}{+0.3 \%}) & \textbf{54.32\%}(\textcolor{green}{+0.92 \%}) & 13130 (\textcolor{red}{+296}) & 41989 (\textcolor{red}{+70}) & \textbf{13.28\%}(\textcolor{green}{-0.48 \%}) & 47.36\%(\textcolor{red}{+1.92 \%})\\
- & APP  & SNIP \cite{lee2018snip}& 4.09 M & 80.4\%(\textcolor{red}{-7.02 \%}) & 41.59\%(\textcolor{red}{-11.81 \%}) & 13939 (\textcolor{red}{+1105}) & 45885 (\textcolor{red}{+3966}) & 20.036\%(\textcolor{red}{+6.276 \%}) & 56.16\%(\textcolor{red}{+10.72 \%}) \\
\hline
ResNet-50 & Baseline & - & 23.5 M & 80.08\% & 42.06\% & 21436& 51866 & 20.64\% & 58.56\% \\
 - & Anytime OSP & SNIP \cite{lee2018snip}& 8.6 M & \textbf{83.95\%}(\textcolor{green}{+3.87 \%}) & \textbf{46.18\%}(\textcolor{green}{+4.12 \%}) & 17521 (\textcolor{green}{-3915}) & \textbf{49083} (\textcolor{green}{-2783}) & 16.942\%(\textcolor{green}{-3.698 \%}) & 56.64\%(\textcolor{green}{-1.92 \%})\\ 
 - & APP & SNIP \cite{lee2018snip}& 8.6 M & 80.86\%(\textcolor{green}{+0.78 \%}) & \textbf{36.78\%}(\textcolor{red}{-5.28 \%}) & \textbf{17073} (\textcolor{green}{-4363}) & 51068 (\textcolor{green}{-798}) & \textbf{20.462\%}(\textcolor{green}{-0.178 \%}) & \textbf{64.213\%}(\textcolor{red}{+5.653 \%}) \\
\hline
\rowcolor{LightCyan}
ResNet-50 & Baseline & - & 23.5 M & 69.95\% & 41.65\% &
25744 & \textbf{50303} &
24.213\% &57.76\%\\
\rowcolor{LightCyan}
- & Baseline (WD) & - & 23.5 M  & 78.05\%(\textcolor{green}{+8.1 \%}) & 41.14\%(\textcolor{red}{-0.51 \%}) & 19925 (\textcolor{green}{-5819}) & 52049 (\textcolor{red}{+1746}) & 23.822\%(\textcolor{green}{-0.391 \%}) & 63.769\%(\textcolor{red}{+6.009 \%}) \\
\rowcolor{LightCyan}
- & Anytime OSP & SNIP \cite{lee2018snip} & 8.6 M & \textbf{71.18\%}(\textcolor{green}{+1.22 \%}) & 38.5(\textcolor{red}{-3.15 \%}) &
25452 (\textcolor{green}{-292}) & \textbf{52953} (\textcolor{red}{+2650}) &
25.084\%(\textcolor{red}{+0.871 \%}) & 56.836\%(\textcolor{green}{-0.924 \%})\\
\rowcolor{LightCyan}
- & APP & SNIP \cite{lee2018snip} & 8.6 M & 67.69\%(\textcolor{red}{-2.26 \%}) & 21.12\%(\textcolor{red}{-20.53 \%}) &
\textbf{23568} (\textcolor{green}{-2176}) & 59042 (\textcolor{red}{+8739}) &
31.236\%(\textcolor{red}{+7.023 \%})& 52.356\%(\textcolor{green}{-5.404 \%})\\
\rowcolor{LightCyan}
- & Anytime OSP & Magnitude & 8.6 M & \textbf{74.48\%}(\textcolor{green}{+4.53 \%}) & \textbf{41.85\%}(\textcolor{green}{+0.2 \%}) &
\textbf{22922} (\textcolor{green}{-2822}) & \textbf{50932} (\textcolor{red}{+629}) &
25.689\%(\textcolor{red}{+1.559 \%})& 58.827\%(\textcolor{red}{+1.067 \%})\\
\rowcolor{LightCyan}
- & APP & Magnitude & 8.6 M & 61.31\%(\textcolor{red}{-8.64 \%}) &20.96\%(\textcolor{red}{-20.69 \%}) &
28152 (\textcolor{red}{+2408}) & 59246 (\textcolor{red}{+8943}) &
\textbf{13.351\%}(\textcolor{green}{-10.862 \%})& 33.956\%(\textcolor{green}{-23.804 \%})\\
\rowcolor{LightCyan}
- & Anytime OSP & Random & 8.6 M & \textbf{70.66\%}(\textcolor{green}{+0.71 \%}) &\textbf{44.08\%}(\textcolor{green}{+2.43 \%}) &
\textbf{25828} (\textcolor{red}{+84}) & \textbf{49006} (\textcolor{green}{-1297}) &
25.778\%(\textcolor{red}{+1.565 \%}) &57.618\%(\textcolor{green}{-0.142 \%})\\
\rowcolor{LightCyan}
- & APP & Random & 8.6 M & 28.01\%(\textcolor{red}{-41.94 \%}) & $1\%^{\dagger}$& 
36044 (\textcolor{red}{+10300}) & 62683 (\textcolor{red}{+12380}) &
${37.742\%}^{\dagger}$ &${25.813\%}^{\dagger}$\\
\rowcolor{LightCyan}
- & Anytime OSP & GraSP \cite{wang2020picking} & 8.6 M & \textbf{80.14\%}(\textcolor{green}{+10.19 \%}) & \textbf{41.56\%}(\textcolor{red}{-0.09 \%}) &
\textbf{18697} (\textcolor{green}{7047}) & \textbf{50867} (\textcolor{red}{+564}) &
\textbf{19.876\%}(\textcolor{green}{-4.337 \%}) & \textbf{53.564\%}(\textcolor{green}{-4.196 \%})\\
\rowcolor{LightCyan}
- & APP & GraSP \cite{wang2020picking} & 8.6 M & $10.0\%^{\dagger}$ & $3.51\%^{\dagger}$ &
41747 (\textcolor{red}{+16003}) & 62672 (\textcolor{red}{+12369}) &
${40.8\%}^{\dagger}$& ${39.378\%}^{\dagger}$\\
\hline
Wide ResNet-50-2 & - & Baseline & 68.9 M & \textbf{84.08\%} & \textbf{47.17\%} & \textbf{17484} &\textbf{ 47725} & \textbf{16.8\%} & \textbf{53.44\%}\\
- & Anytime OSP & SNIP \cite{lee2018snip} & 25.2 M & 81.43\%(\textcolor{red}{-2.65 \%}) & 42.58\%(\textcolor{red}{-4.59 \%}) & 18715 (\textcolor{red}{+1231}) & 52184 (\textcolor{red}{+4459}) & 18.382\% (\textcolor{red}{+1.582 \%}) & 57.28\%(\textcolor{red}{+3.84 \%}) \\
- & APP & SNIP \cite{lee2018snip} & 25.2 M & 81.46\%(\textcolor{red}{-2.62 \%}) & 35.49\%(\textcolor{red}{-11.68}\%) & 18222 (\textcolor{red}{+738}) &51062(\textcolor{red}{+3337}) & 18.72\%(\textcolor{red}{+1.92 \%}) &64.587\%(\textcolor{red}{+11.147 \%}) \\
\hline
VGG-16-BN & Baseline & - & 138.42 M & \textbf{88.1\%} & \textbf{52.24\%} & \textbf{11796} &\textbf{41873} & 15.022\% & \textbf{48\%} \\
- & Anytime OSP & SNIP \cite{lee2018snip} & 50.6 M & 88.01\%(\textcolor{red}{-0.09 \%}) & 50.63\%(\textcolor{red}{-1.61 \%}) & 11874 (\textcolor{red}{+78}) & 42660 (\textcolor{red}{+787}) & \textbf{13.084\%}(\textcolor{green}{-1.938 \%}) & 52.16\%(\textcolor{red}{+4.16 \%})\\
- & APP & SNIP \cite{lee2018snip} & 50.6 M & 81.74\%(\textcolor{red}{-6.36 \%}) & 41.62\%(\textcolor{red}{-10.62 \%}) & 13598 (\textcolor{red}{+1802}) & 44935 (\textcolor{red}{+3062})& 19.449\%(\textcolor{red}{+4.427 \%}) & 61.991\%(\textcolor{red}{+13.991 \%}) \\
\hline
\end{tabular}}
\end{center}
\end{table}
\setlength{\tabcolsep}{1.4pt}

In Table \ref{table:cifar_sgd_cyclic_norep}, we conducted experiments to validate the effect of no replay for APP, Anytime OSP, and the baseline models for different backbones. From the experiments, we can conclude with high certainty that APP requires full replay of megabatches to provide a performance improvement. As shown in the table, we see that APP models cause a significant decrease in performance, while Anytime OSP models improve performance compared to their baseline counterparts. We hypothesize that the loss in performance is induced by the model restructuring caused by pruning at the start of each megabatch, which can be attributed to the loss in knowledge transfer while transitioning from one megabatch to the next. 

\subsubsection{Analysis of moderate and long sequence ALMA ($|S_B| = 25, 50, 100$)}
For validation on variation of $|S_B|$, we conducted experiments using only the ResNet-50 model with full replay and with SNIP as the pruner of choice for both APP and Anytime OSP variants as reported in Table \ref{table:cifar_large}. Similarly to short-sequence-based ALMA, we observed a strong improvement in performance while using APP compared to the Anytime OSP and baseline models. In particular, when $|S_B| = 100$, where each megabatch has $|\mathcal{M}_t| = 500$ samples, we report an improvement in CER by 105277 compared to the baseline model, which is equivalent to APP correctly classifying the test set $T_{x,y}$ of 10,000 samples 10 times compared to the baseline model throughout the training process on the complete stream $|S_B|$. Interestingly, we find that the performance of the baseline has a high variation caused by the change in $|S_B|$ with a deviation in test accuracy of $\sigma = 4.345\%$, while APP is extremely stable and is less sensitive to the change in $|S_B|$ with a deviation in test accuracy of $\sigma = 1.743\%$ across $|S_B|$ values of 8, 25, 50 and 100.
We further analyze and investigate the training dynamics observed during training moderate- and long-sequence ALMA, which we discuss in detail in Section \ref{subsec: phase}.

\setlength{\tabcolsep}{4pt}
\begin{table}[H]
\begin{center}
\caption{Results of ALMA training of C-10 models with varying $|S_B|$.}
\label{table:cifar_large}
\resizebox{\textwidth}{!}{\begin{tabular}{ccccccccc}
\hline\noalign{\smallskip}
Backbone & Method & Pruner & $|\theta|$($\downarrow$) & $|S_B|$ & $|\mathcal{M}_t|$ & Test Accuracy($\uparrow$) & CER ($\downarrow$) & Generalization Gap($\downarrow$)\\
\noalign{\smallskip}
\hline
\noalign{\smallskip}
ResNet-50 & Baseline & - & 23.5 M & 25 & 2000 & \textbf{82.69\%} & 118876 & 9.978\% \\
- & Anytime OSP & SNIP \cite{lee2018snip} & 8.6 M & 25 & 2000 & 78.86\%(\textcolor{red}{-3.83 \%})& 110698 (\textcolor{green}{-8178}) & 16.284\%(\textcolor{red}{+6.306 \%}) \\
- & APP & SNIP \cite{lee2018snip} & 8.6 M & 25 & 2000 & 79.73\%(\textcolor{red}{-2.96 \%})& \textbf{104435} (\textcolor{green}{-14441}) & \textbf{2.916\%}(\textcolor{green}{-7.062 \%})\\
\hline
ResNet-50 & Baseline & - & 23.5 M & 50 & 1000 & 79.13\% & 193384 & 20.971\% \\
- & Anytime OSP & SNIP \cite{lee2018snip} & 8.6 M & 50 & 1000 & 72.91\%(\textcolor{red}{-6.22 \%})& 202212 (\textcolor{red}{+8828}) & 26.56\%(\textcolor{green}{-2.411 \%}) \\
- & APP & SNIP \cite{lee2018snip} & 8.6 M & 50 & 1000 & \textbf{82.0\%}(\textcolor{green}{+2.87 \%})& \textbf{163503} (\textcolor{green}{-29881}) & \textbf{14.707\%}(\textcolor{green}{-6.264 \%})\\
\hline
ResNet-50 & Baseline & - & 23.5 M & 100 & 500 & 70.87\% & 396572 & 28.971\% \\
- & Anytime OSP & SNIP \cite{lee2018snip} & 8.6 M & 100 & 500 & 78.51\%(\textcolor{green}{+7.64 \%}) & 315349 (\textcolor{green}{-81223}) & 20.133\%(\textcolor{green}{-8.838 \%}) \\
- & APP & SNIP \cite{lee2018snip} & 8.6 M & 100 & 500 & \textbf{82.32\%}(\textcolor{green}{+11.45 \%}) & \textbf{291295} (\textcolor{green}{-105277}) & \textbf{16.502\%}(\textcolor{green}{-12.469 \%})\\
\hline
\end{tabular}}
\end{center}
\end{table}
\setlength{\tabcolsep}{1.4pt}

\subsubsection{Few shot experiments on Restricted ImageNet}

\begin{figure}[H]
\centering
\includegraphics[width=0.98\textwidth]{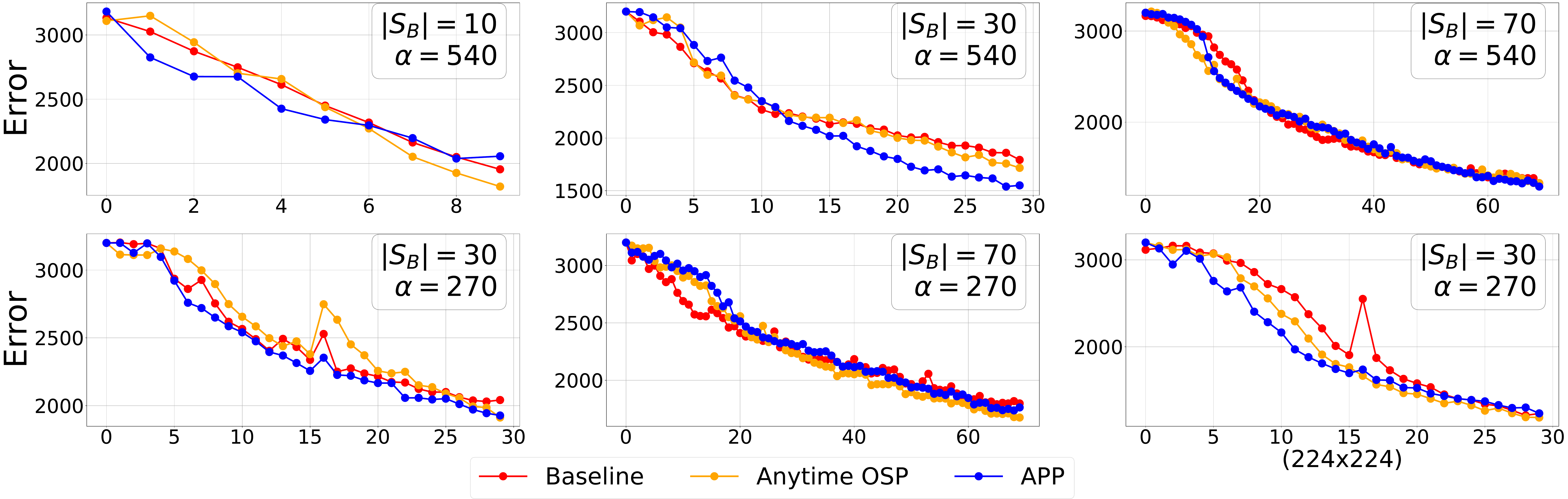}
\caption{Error rates on the test set of the trained model at each megabatch $\mathcal{M}_t$ for the restricted Imagenet experiments reported in Table \ref{table:im32_few}.}
\label{fig:CER}
\end{figure}

\begin{figure}[H]
\centering
\includegraphics[width=0.98\textwidth]{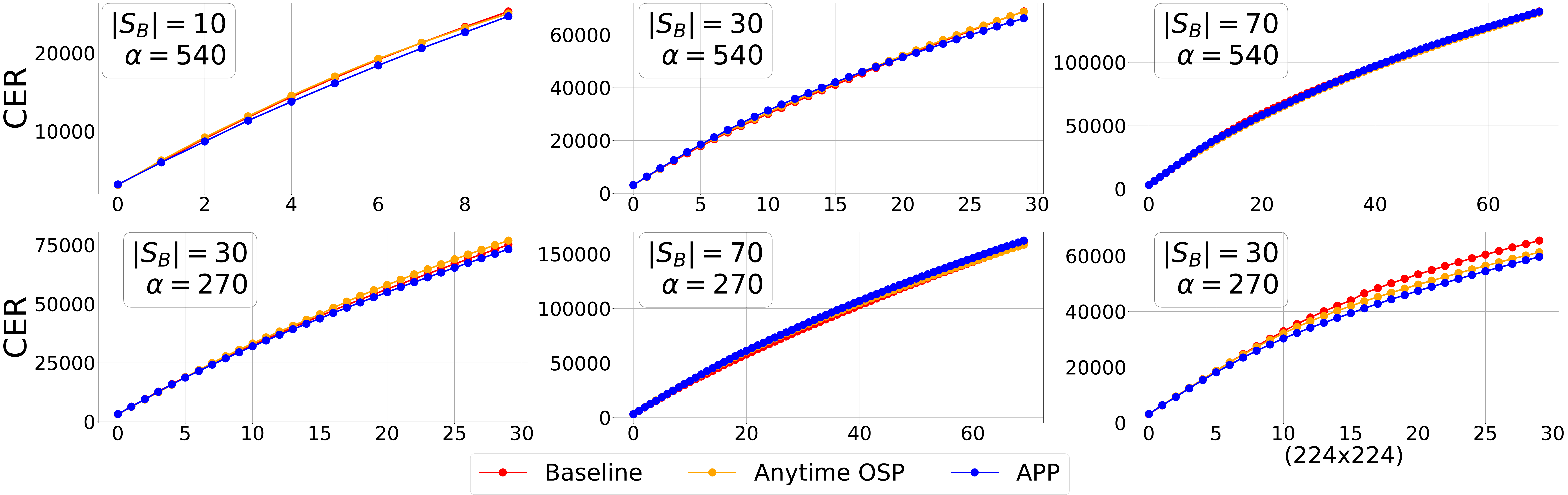}
\caption{CER of the trained model at each megabatch $\mathcal{M}_t$ for the restricted Imagenet experiments reported in Table \ref{table:im32_few}.}
\label{fig:CER_rate}
\end{figure}

In this section, we investigate the performance of APP compared to Anytime OSP and the baseline models on Restricted Balanced ImageNet \cite{robustness,tsipras2018robustness} using various few-shot learning settings. We primarily conduct experiments using the following two few-shot settings.
\begin{enumerate}
    \item \textbf{$\alpha = 270$}: For this, we only keep 270 samples per class in the complete dataset, which essentially totals 3780 samples for the complete dataset. We tested this using two different number of megabatches $|S_B| = 30, 70$ such that each megabatch consists of $|\mathcal{M}_t| = 126, 54$ samples, respectively. For $|S_B| = 30$, we performed experiments on the 224 x 224 and 32 x 32 sizes of the data set. For $|S_B| = 30$, we reduce the minibatch size of each megabatch $\mathcal{M}_t$ to 64 while for $|S_B| = 70$, we reduce it to 32. 
    \item \textbf{$\alpha = 540$}: For this, we only keep 540 samples per class in the complete dataset which essentially totals 7560 samples for the complete dataset. We test this using three different number of megabatches $|S_B| = 10, 30, 70$ such that each megabatch consists of $|\mathcal{M}_t| = 756, 252, 108$ samples, respectively. For $|S_B| = 70$, we reduce the mini-batch size of each megabatch $\mathcal{M}_t$ to 64.
\end{enumerate}

As reported in Table \ref{table:im32_few}, we observe that APP significantly reduces the generalization gap for each model variant compared to the Anytime OSP and baseline counterparts. Excluding $\alpha = 270$ with $|S_B| = 70$ experiment on the 32 x 32 downsampled version of restricted ImageNet, we observe a decrease in CER compared to the baseline model. For example, for $\alpha = 270$ with $|S_B| = 30$ on the 224 x 224 version of Restricted Imagenet, we observe that APP reduces the CER by 5846 compared to baseline, which essentially means that APP correctly classified $\approx 1.5\times$ the test set throughout the training on the full stream $S_B$. We also observe strong notable improvements in test accuracy for anytime OSP models in the $\alpha = 270$ setting, where it records the highest test accuracy in all experiments. 

We also visualize and compare the error rate on the test set and CER for each megabatch for APP, Anytime OSP and baseline models in Fig. \ref{fig:CER} and Fig. \ref{fig:CER_rate} respectively. We observe that while the final CER for APP with $\alpha = 270$ and $|S_B| = 70$ is higher than the baseline, this is caused by the higher error rates at the initial megabatches for APP as shown in the fifth subplot (2\textsuperscript{nd} row, 2\textsuperscript{nd} column) of Fig. \ref{fig:CER}, while APP at the final megabatches had a lower error than the baseline. In both figures, we observe that APP consistently retains both lower error and CER in almost every megabatch in all settings reported in Table \ref{table:im32_few}. 

\setlength{\tabcolsep}{4pt}
\begin{table}[H]
\begin{center}
\caption{Results on Few-shot ImageNet Restricted ALMA using SGD with cyclic multistep decay at every $\mathcal{M}_t$.}
\label{table:im32_few}
\resizebox{\textwidth}{!}{\begin{tabular}{ccccccccccc}
\hline\noalign{\smallskip}
Method & Resolution & $|S_B|$  &$|\mathcal{M}_t|$ & $\alpha$ &  Test Accuracy($\uparrow$) &
CER($\downarrow$) &
Generalization Gap($\downarrow$)\\
\noalign{\smallskip}
\hline
\noalign{\smallskip}
Baseline & 32 x 32 & 10  & 756 & 540 & 43.36\% & 25328 & 17.394\%\\
Anytime OSP & - & - & - & - & \textbf{47.246\%}(\textcolor{green}{+3.886 \%}) & 24978 (\textcolor{green}{-350}) & 21.529\%(\textcolor{red}{+4.135 \%}) \\
APP & - & - & - & - & 40.40\%(\textcolor{red}{-2.96 \%}) & \textbf{24712} (\textcolor{green}{-616}) & \textbf{6.963\%}(\textcolor{green}{-10.431 \%}) \\
\hline
Baseline & - & 30 & 126 & 270 & 40.811\% & 75128 & 55.503\%\\
Anytime OSP & - & - & - & - & \textbf{44.55\%}(\textcolor{green}{+3.739 \%}) & 76871 (\textcolor{red}{+1743}) & 48.53\%(\textcolor{green}{-6.973 \%}) \\
APP & - & - & - & - & 44.11\%(\textcolor{green}{+3.229 \%}) & \textbf{73206} (\textcolor{green}{-1922}) & \textbf{34.423\%}(\textcolor{green}{-21.08 \%}) \\
\hline
Baseline & - & 30 & 252 & 540 & 48.03\% & 68832 & 48.733\%\\
Anytime OSP & - & - & - & - & 50.23\%(\textcolor{green}{+2.2 \%}) & 68765 (\textcolor{green}{-67}) & 45.288\%(\textcolor{green}{-3.445 \%}) \\
APP & - & - & - & - & \textbf{55.04\%}(\textcolor{green}{+7.01 \%}) & \textbf{66239} (\textcolor{green}{-2593}) & \textbf{26.388\%}(\textcolor{green}{-22.345 \%}) \\
\hline
Baseline & - & 70 & 54 & 270 & 47.88\% & 159204 & 45.03\%\\
Anytime OSP & - & - & - & - & \textbf{51.449\%}(\textcolor{green}{+3.569 \%}) & \textbf{158608} (\textcolor{green}{-596}) & 45.357\%(\textcolor{red}{+0.327 \%}) \\
APP & - & - & - & - & 48.898\%(\textcolor{green}{+1.018 \%}) & 162360 (\textcolor{red}{+3156}) & \textbf{30.744\%}(\textcolor{green}{-14.286 \%}) \\
\hline

Baseline & - & 70 & 108 & 540 & 61.391\% & 140069 & 34.456\%\\
Anytime OSP & - & - & - & - & 61.391\%(\textcolor{gray}{0\%}) & \textbf{139152} (\textcolor{green}{-917}) & 32.979\%(\textcolor{green}{-1.477 \%}) \\
APP & - & - & - & - & \textbf{62.492\%}(\textcolor{green}{+1.101 \%}) & 139963 (\textcolor{green}{-106}) & \textbf{17.5859\%}(\textcolor{green}{-16.8701 \%}) \\
\hline
\hline
Baseline & 224 x 224 & 30 & 126 & 270 & 64.289\% & 65525 & 32.149\%\\
Anytime OSP & - & - & - & - & \textbf{65.623\%}(\textcolor{green}{+1.334 \%}) & 61341 (\textcolor{green}{-4184}) & 33.435\%(\textcolor{red}{+1.286 \%}) \\
APP & - & - & - & - & 64.231\%(\textcolor{red}{-0.058 \%}) & \textbf{59679} (\textcolor{green}{-5846}) & \textbf{29.884\%}(\textcolor{green}{-2.265 \%}) \\
\hline
\end{tabular}}
\end{center}
\end{table}
\setlength{\tabcolsep}{1.4pt}

\subsubsection{Analysis of training curves and CER for C-10/100}

\begin{figure}[t]
\centering
  \subcaptionbox{Best validation accuracy achieved at each $\mathcal{M}_t$ for APP with various pruners on a ResNet-50 backbone for C-10,100 with $|S_B|=8$ and $\tau = 4.5$.\label{fig:train(a)}}{\includegraphics[width=0.31\linewidth]{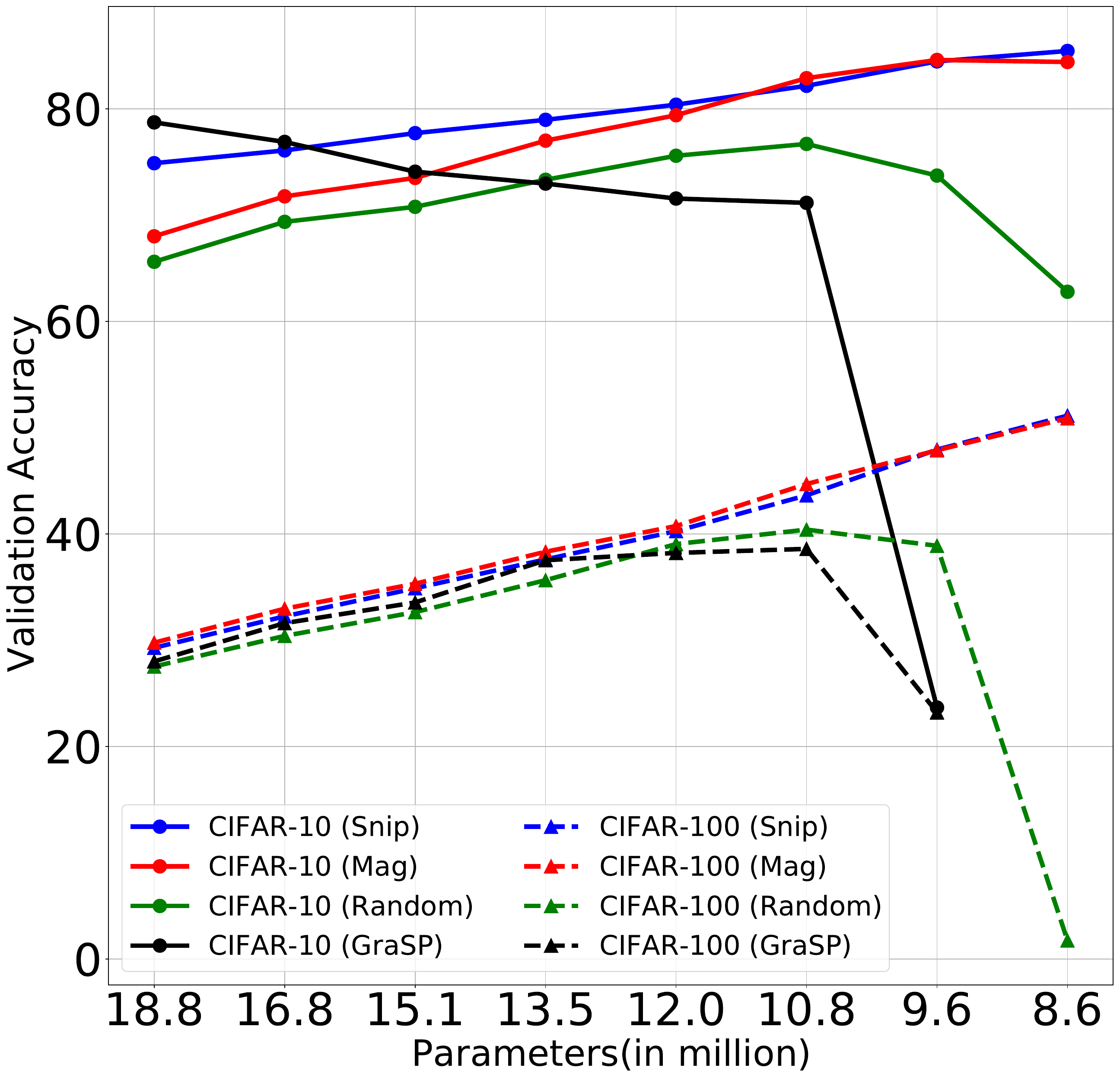}}\hspace{1em}%
  \subcaptionbox{Comparison of train accuracy curves \textcolor{blue}{APP}(SNIP), \textcolor{orange}{Anytime OSP}(SNIP) and \textcolor{red}{Baseline} with a ResNet-50 backbone trained with $|S_B|=50$ on C-10.\label{fig:train(b)}}{\includegraphics[width=0.31\linewidth]{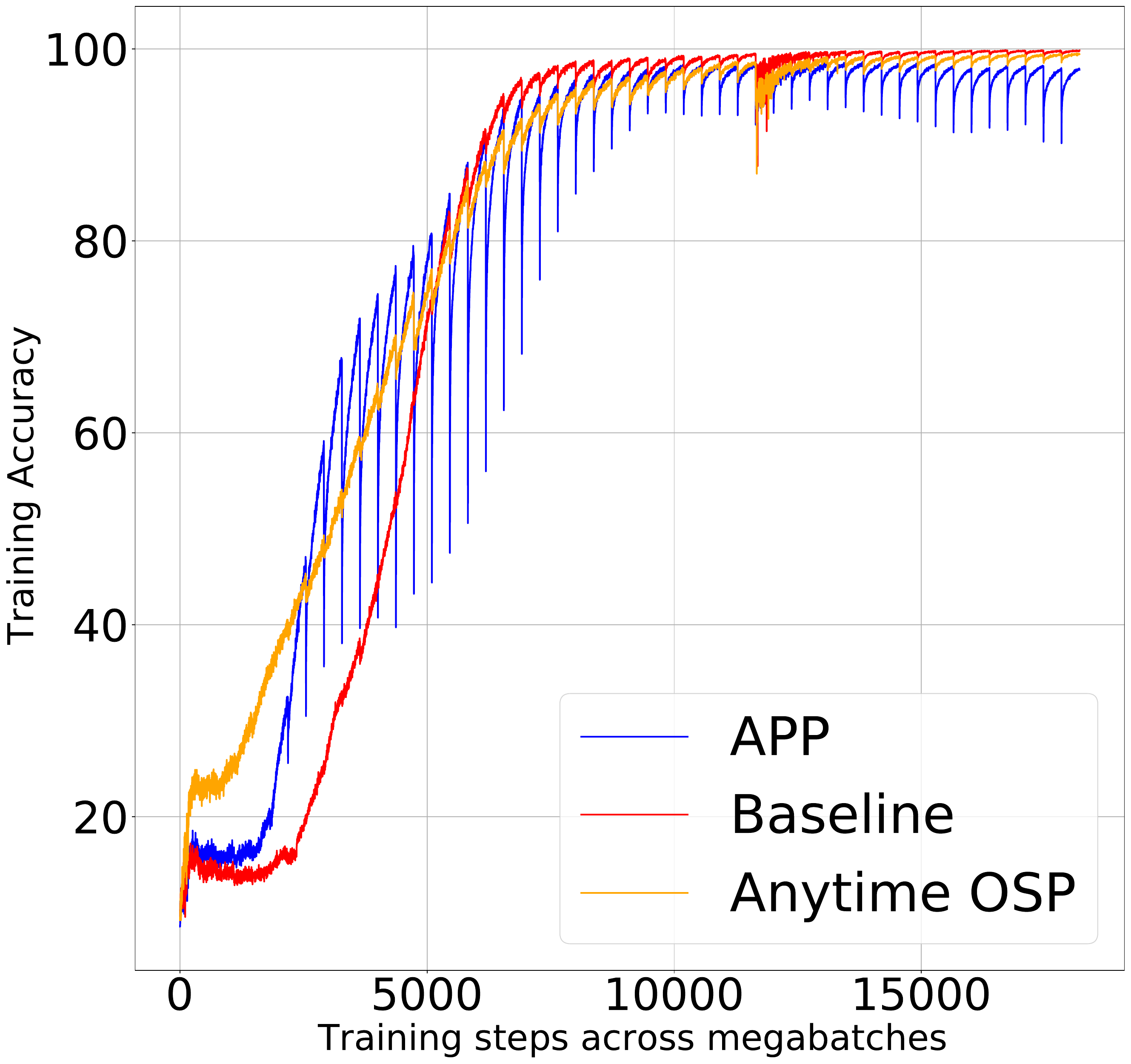}}\hspace{1em}%
  \subcaptionbox{Comparison of validation loss curves  \textcolor{blue}{APP} (SNIP), \textcolor{orange}{Anytime OSP} (SNIP) and \textcolor{red}{Baseline} with a ResNet-50 backbone trained with $|S_B|=50$ on C-10.\label{fig:train(c)}}{\includegraphics[width=0.31\linewidth]{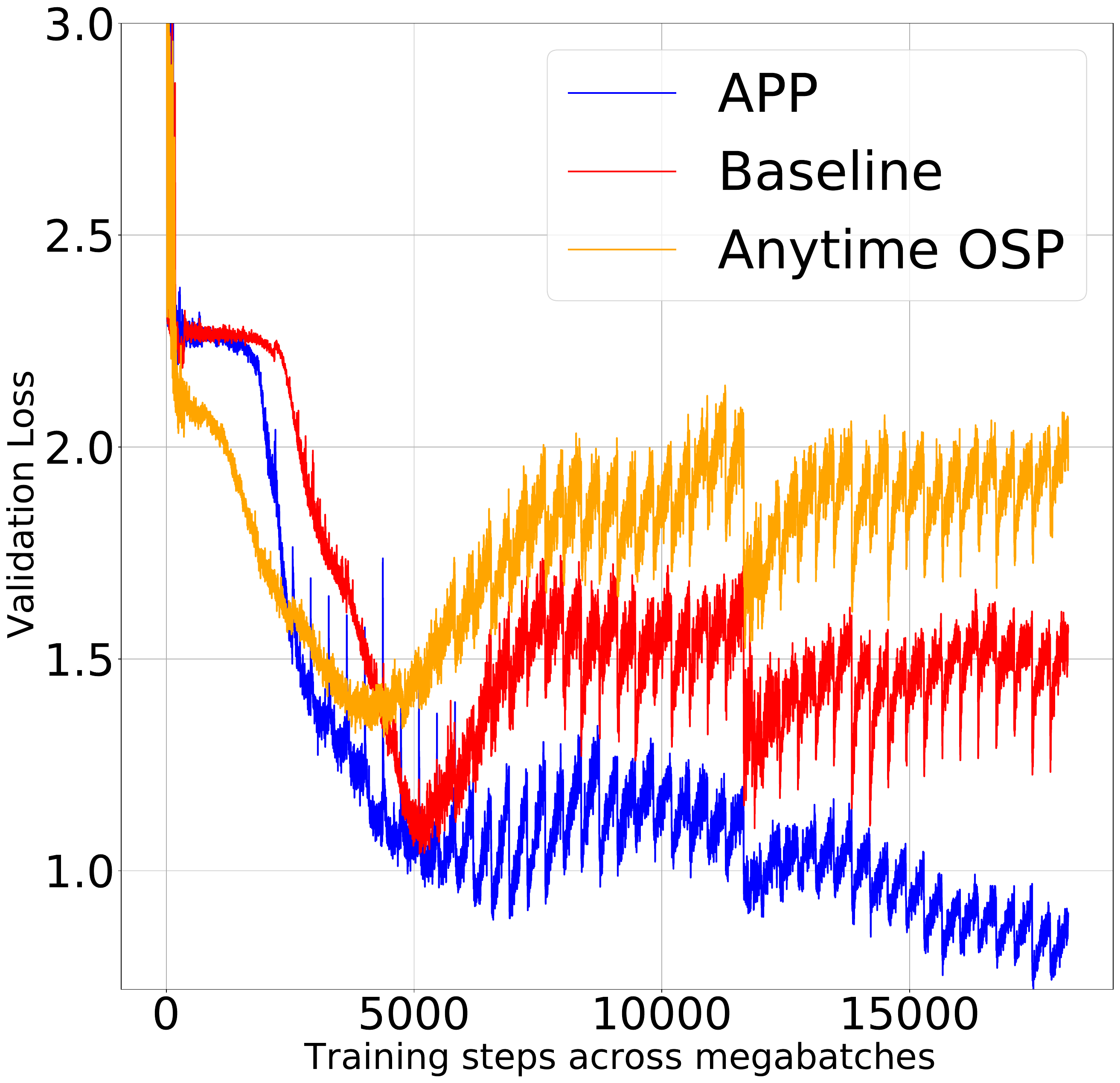}}
\label{fig:train}
\end{figure}

In Fig. \ref{fig:train(b)} and Fig. \ref{fig:train(c)}, we start by analyzing the learning curves, specifically the training accuracy and validation loss curves on C-10 for APP, Anytime OSP, and the baseline models as a function of the total number of training iterations on the entire stream of megabatches $|S_B|$. First, in Fig. \ref{fig:train(b)}, we observe a distinct oscillation in the training accuracy curve for APP, which is caused by pruning at the start of training on each new megabatch $\mathcal{M}_t$, resulting in a sharp drop in the initial point accuracy. Second, we also observe in Fig. \ref{fig:train(c)}, that the validation loss curve for APP has a negative slope while approaching the completion of training over the complete stream $S_B$, while the curves for Anytime OSP and baseline models are significantly higher and plateauing, indicating saturation in learning capacity.

Furthermore, we also visualize the best validation accuracy achieved for APP with various pruners on a ResNet-50 backbone for C-10,100 with $|S_B|=8$ and $\tau = 4.5$. We observe that SNIP and magnitude-based pruning provide consistent and stable performance improvements over each megabatch in the stream $S_B$, while random pruning and GraSP cause instability and drop performance by a significant margin during training on the final megabatches in the stream $S_B$. Thus, we set SNIP to be the pruner of choice for APP by default for all of our experiments. 

\begin{figure}[t]
\centering
\includegraphics[width=0.98\textwidth]{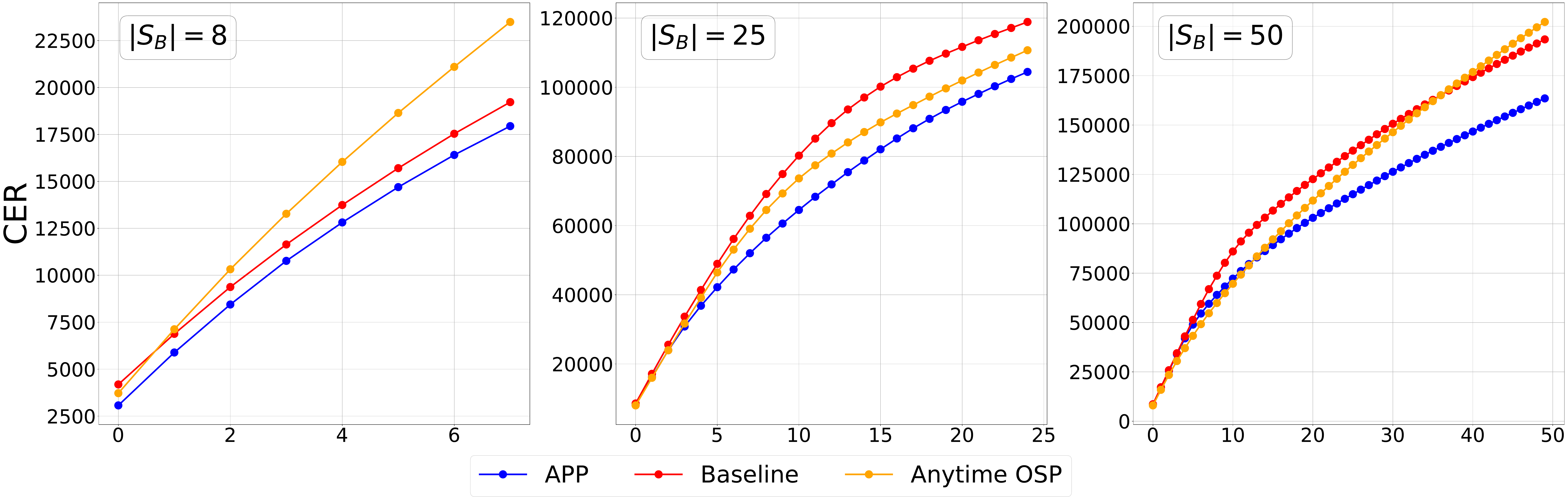}
\caption{Change in CER during training of \textcolor{blue}{APP} (SNIP), \textcolor{orange}{Anytime OSP} (SNIP) and \textcolor{red}{Baseline} using a ResNet-50 on C-10 with varying number of megabatches($|S_B|$). X-axis represents the number of megabatches in the entire stream $S_B$.}
\label{fig: cer}
\end{figure}

Finally, we visualize the change in CER during training for APP (SNIP), Anytime OSP (SNIP) and baseline using a ResNet-50 on C-10 by varying the total number of megabatches ($|S_B|$) in the stream $S_B$. As shown in Fig. \ref{fig: cer}, APP (SNIP) consistently maintains a lower CER compared to its Anytime OSP and baseline counterparts under the short ($|S_B| = 8$), moderate ($|S_B| = 25$) and long ($|S_B| = 50$) ALMA sequence.

\subsubsection{Restricted ImageNet full ALMA}

Finally, we also conducted an experiment on the full restricted ImageNet Balanced dataset (32 x 32 downsampled version) using $|S_B| = 3$ megabatches with each megabatch containing $\mathcal{M}_t = 29839$ samples on a ResNet-50 trained using SGD and cyclic multidecay learning rate policy at each megabatch $\mathcal{M}_t$. As reported in Table \ref{table:im32}, we observe that both APP and Anytime OSP models cause a drop in test accuracy and an increase in CER compared to the baseline model. However, APP reduces the generalization gap by a margin of $1.845\%$ for $|S_B| = 3$ and $6.077\%$ for $|S_B| = 53$ compared to the baseline.

\setlength{\tabcolsep}{4pt}
\begin{table}[H]
\begin{center}
\caption{Results on Restricted ImageNet (32 x 32) ALMA using SGD with cyclic multi-step decay at every $\mathcal{M}_t$.}
\label{table:im32}
\resizebox{\textwidth}{!}{\begin{tabular}{ccccccccc}
\hline\noalign{\smallskip}
Method &$|S_B|$ & $|\mathcal{M}_t|$ & Test Accuracy($\uparrow$) &
CER($\downarrow$) &
Generalization Gap($\downarrow$)\\
\noalign{\smallskip}
\hline
\noalign{\smallskip}
Baseline  & 3 & 29839 & \textbf{86.318\%} & \textbf{2046} & 10.792\%\\
Anytime OSP & - & - & 84.55\%(\textcolor{red}{-1.768 \%}) & 2384 (\textcolor{red}{+338}) & 10.095\%(\textcolor{green}{-0.697 \%}) \\
APP & - & - & 84.49\%(\textcolor{red}{-1.828 \%}) & 2310 (\textcolor{red}{+264}) & \textbf{8.947\%}(\textcolor{green}{-1.845 \%}) \\
\hline
Baseline  & 53 & 1689 & \textbf{86.782\%} & \textbf{44702} & 8.546\%\\
Anytime OSP & - & - & 86.492\%(\textcolor{red}{-0.29 \%}) & 47431 (\textcolor{red}{+2729}) & 7.179\%(\textcolor{green}{-1.367 \%}) \\
APP & - & - & 83.333\%(\textcolor{red}{-3.449 \%}) & 51728 (\textcolor{red}{+7026}) & \textbf{2.469\%}(\textcolor{green}{-6.077 \%}) \\
\hline
\end{tabular}}
\end{center}
\end{table}
\setlength{\tabcolsep}{1.4pt}

\subsection{Transitions in generalization gap}
\label{subsec: phase}

\begin{figure}[H]
\centering
\includegraphics[width=0.98\textwidth]{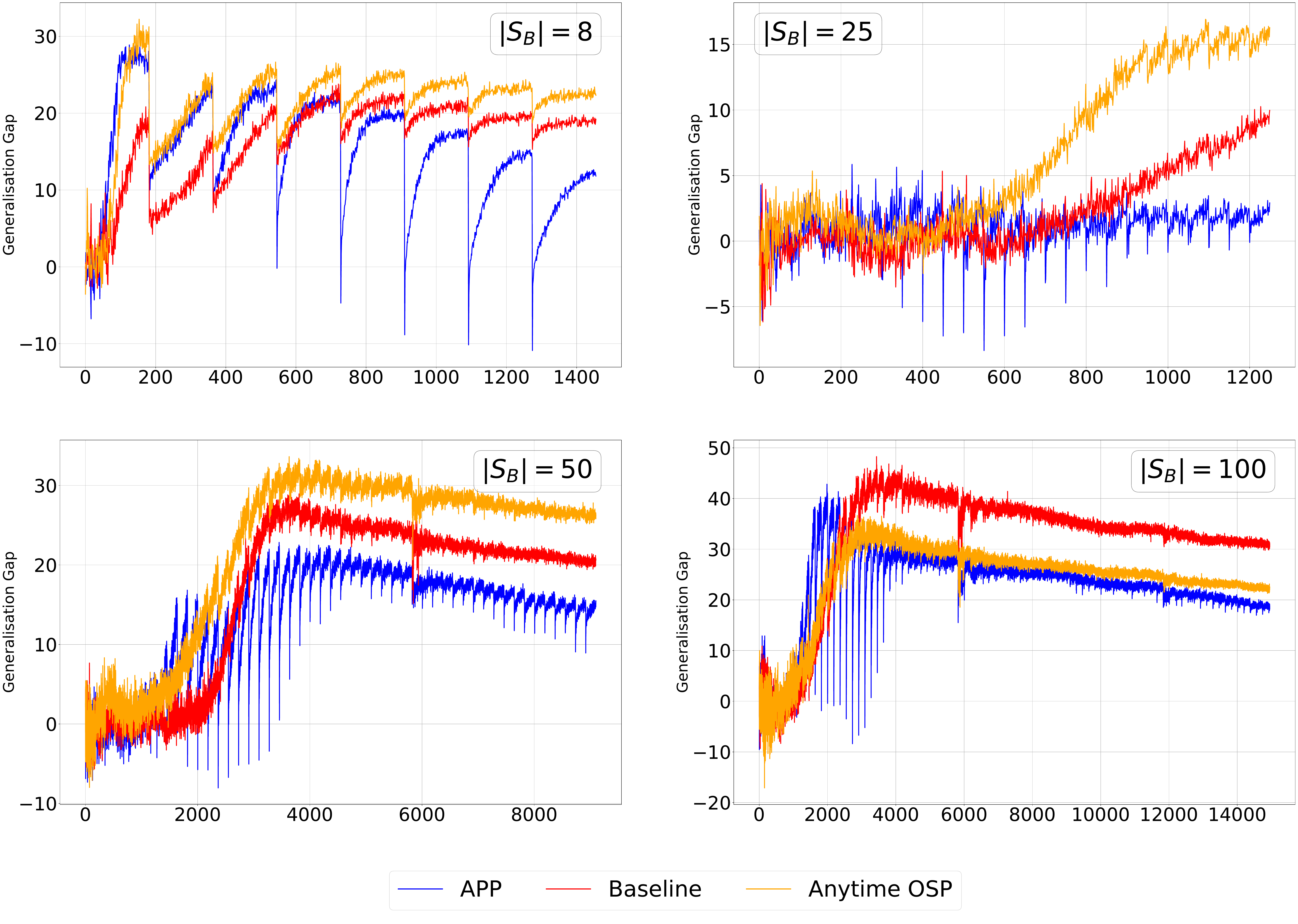}
\caption{Generalization gap curves during training of \textcolor{blue}{APP (SNIP)}, \textcolor{orange}{Anytime OSP (SNIP)} and \textcolor{red}{Baseline} using a ResNet-50 on C-10 with varying number of megabatches($|S_B|$) with full replay. X-axis represents the total number of training iterations completed throughout the stream $S_B$.}
\label{fig:gen}
\end{figure}

\begin{figure}[t]
\centering
\includegraphics[width=0.98\textwidth]{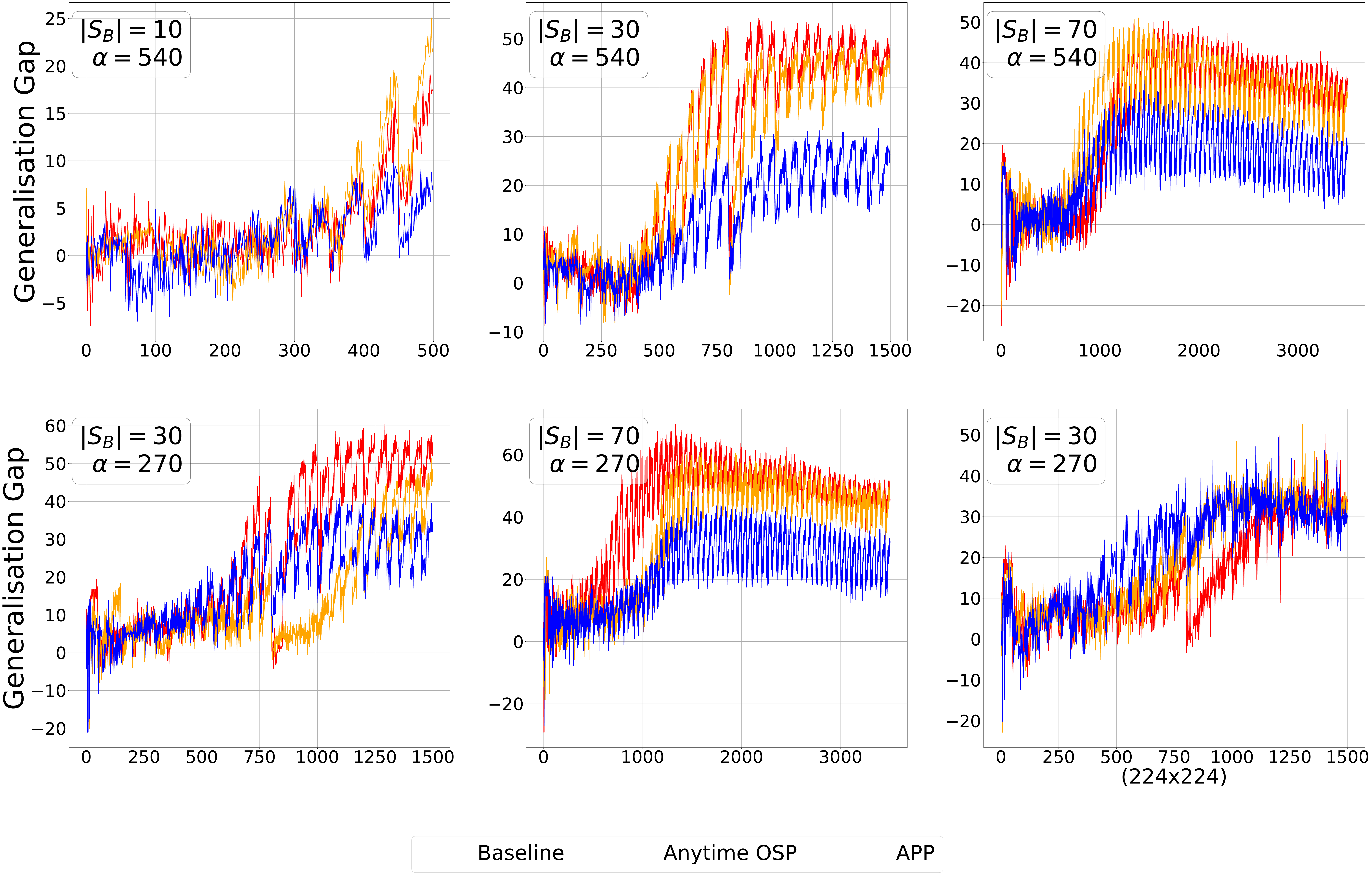}
\caption{Generalization gap curves during training of \textcolor{blue}{APP (SNIP)}, \textcolor{orange}{Anytime OSP (SNIP)} and \textcolor{red}{Baseline} for the experiments reported in Table \ref{table:im32_few}.}
\label{fig:gen_img}
\end{figure}

While training the models for empirical validation, we observed a very interesting trend in the training dynamics, precisely the generalization gap, in the long-sequence ALMA. As defined in Section \ref{sec: app}, the generalization gap is the difference observed between the training and validation accuracy across the complete training process over the stream $S_B$. The generalization gap is used to conclude whether a model is overfitting or underfitting, thus serving as an important criterion for the evaluation of models and investigating failure modes during model training. Similarly to the results reported in \cite{nakkiran2021deep}, we observe a non-monotonic transition in the generalization gap across APP, Anytime OSP, and baseline models during long-sequence ALMA training ($|S_B| = (50,100)$). In Fig. \ref{fig:gen}, we observe the generalization gap as a function of training iterations over the entire stream $S_B$ for APP, Anytime OSP and baseline models using ResNet-50 backbone on C-10 with various number of megabatches ($|S_B| = (8,25,50,100)$). We observe in both $|S_B| = 50$ and $|S_B| = 100$, a non-monotonic transition in the generalization gap where the model starts by underfitting, then sharply goes into the critical regime of overfitting, and subsequently has a gradual decrease in the generalization gap. We additionally observe that the generalization gap curve for APP tends to oscillate heavily in the critical regime, which might be attributed to pruning at the start of the megabatch under fewer data scenarios. Additionally, we also observe that for $|S_B| = 25$, the generalization gap for the Anytime OSP and the baseline model, rises sharply towards the end of training, while for APP it remains relatively stable.  

We also visualize the generalization gap as a function of training iterations in Fig. \ref{fig:gen_img} for the experiments reported in Table \ref{table:im32_few}. As demonstrated in Fig. \ref{fig:gen}, we observe the same non-monotonic transition in the high number of megabatch $|S_B| = 30,70$ settings. In all subplots, it can be seen that APP consistently maintains a lower generalization gap compared to its Anytime OSP and baseline counterparts. 

\subsection{Layer-wise Pruning Distribution}

In this section, we analyze the distribution of the pruned weights across the layers when using different pruners of choice for APP. In the experiment, we only visualize the difference between magnitude-based pruning and SNIP \cite{lee2018snip}, since random pruning and GraSP \cite{wang2020picking} lead to unstable training and therefore do not provide any meaningful insight. For the backbone, we used a ResNet-50 with an SGD + multidecay learning rate policy for the first megabatch $\mathcal{M}_1$ only. Both models were trained with full replay for a total of $|S_B| = 8$ megabatches, each megabatch $\mathcal{M}_t$ having a total of $|\mathcal{M}_t| = 6250$ samples. 

\begin{figure}[H]
\centering
\includegraphics[width=0.98\textwidth]{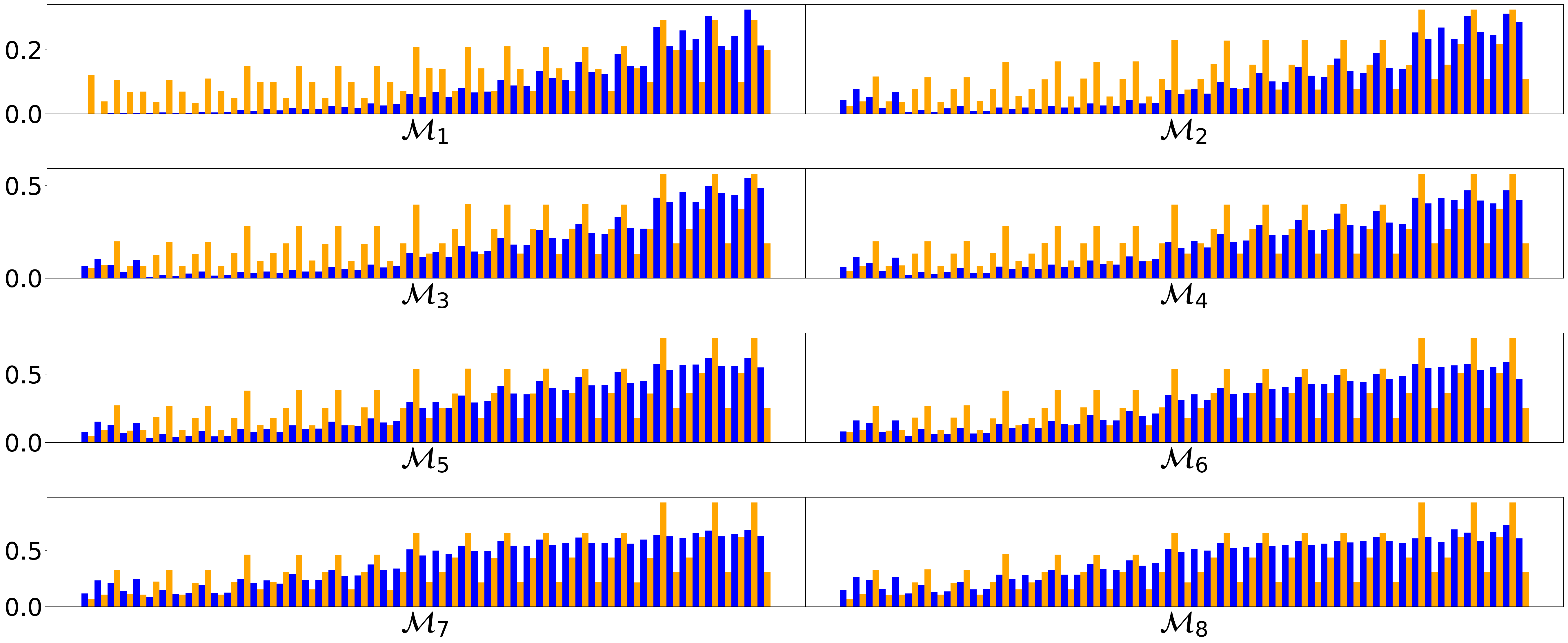}
\caption{Progressive number of weights pruned at each megabatch $\mathcal{M}_t$ for ALMA on ResNet-50 with APP for a total of $|S_B| = 8$ megabatches on the CIFAR-10 dataset with full replay. For the experiment, we used the SGD + cyclic learning rate policy at only the first megabatch $\mathcal{M}_1$. \textcolor{blue}{Bars} represent APP with SNIP as the pruner of choice while \textcolor{orange}{Bars} represent APP with magnitude pruning. Both pruners were used for the same target sparsity $\tau = 4.5$. In each subplot, each bar corresponds to each layer of the network, and the y-axis represents the \% of the weights pruned for that layer.}
\label{fig:dist}
\end{figure}

As demonstrated in Fig. \ref{fig:dist}, we see that magnitude pruning leads to more weights of the initial layers being pruned at the initial megabatches compared to SNIP. \cite{shang2016understanding,xiao2021early} have demonstrated the importance of early convolution layers in the performance of deep convolution neural networks, and it is a well-accepted notion that early convolution layers are responsible for learning low-level features, such as edges, while later layers learn high-level features, such as texture. Since magnitude-based pruning removes a significant amount of early layer weights, this causes a drop in test accuracy compared to SNIP, which prunes more of the latter layers at the initial megabatches. 

\section{Reproducibility Statement}

To ensure fair and reproducible experiments throughout our work, we enforced the following measures: 
\begin{enumerate}
    \item \textbf{Use of publicly available open source datasets}: As defined in subsection \ref{sec:data}, throughout our research, we do empirical evaluation only using publicly available datasets - (a) CIFAR-10 \cite{krizhevsky2009learning}, (b) CIFAR-100 \cite{krizhevsky2009learning}, and (c) Restricted ImageNet \cite{tsipras2018robustness}. In our code, we also provide predefined dataloaders and augmentations that were used to construct the megabatches $\mathcal{M}_t$. None of the datasets used in this work contain sensitive or private information pertaining to an individual or a single entity against their consent.
    \item \textbf{Use of open source frameworks and packages}: For all empirical experiments, we rely on packages and libraries that are accessible and available to the general public. 
\end{enumerate}

\subsection{Hardware resources}

For all experiments, we primarily used three different hardware configurations, as listed below:
\begin{enumerate}
    \item 1 NVIDIA A100 GPU accelerator with 20 CPUs and 24 GB memory.
    \item 1 NVIDIA V100 GPU accelerator with 20 CPUs and 18 GB memory.
    \item 1 NVIDIA RTX 8000 GPU accelerator with 8,20 CPUs and 12 GB memory.
\end{enumerate}

All CIFAR-10 and CIFAR-100 experiments were conducted using the NVIDIA RTX-8000 GPU, while the NVIDIA A100 and V100 were only used for restricted ImageNet experiments. Finally, we also used Google Colaboratory for initial proof-of-concept and ablation experiments.

\section{Conclusion, Open Questions and Future Work}

In this work, we introduced Anytime Progressive Pruning (APP), a novel way to progressively prune deep networks while training in an ALMA regime. We improvise on existing pruning at initialization strategies to design APP and perform an extensive empirical evaluation to validate performance improvement in various architectures and datasets. We found that pruning deep networks with APP while training in an ALMA setting causes a significant drop in the generalization gap compared to one-shot pruning methods and the dense baseline model. 

We conclude this research with the remark that our work serves to lay the foundation for further exploration into dynamic and progressive pruning in sequential learning regimes. Although our work provides constructive insights into the training dynamics of progressive pruning, there are several questions that we hope can be subsequently explored based on this work, which are as follows.

\begin{enumerate}
    \item How can we control the pruning rate at each megabatch $\mathcal{M}_t$ without prior knowledge of the total number of megabatches in the stream $S_B$?
    \item What is the reason behind the non-monotonic transitions observed in the generalisation gap?
    \item Although we hypothesize that the reason behind the oscillation (drop in test accuracy at the initial iteration of each megabatch) for APP is due to the regularization effect induced due to pruning, how can we formally prove this phenomenon?
    \item Why does APP not work under no replay settings, while Anytime OSP does?
\end{enumerate}

In addition to the above questions, in future work, we aim to further improve the performance of APP in no-replay settings by designing an optimal framework for data-constrained progressive pruning. We also aim to improve the performance of APP for greater target sparsity $\tau$ and simultaneously perform a hyperparameter search to find the optimal hyperparameters for progressive pruning using APP. Finally, we also aim to transfer the progressive pruning setting to other tasks such as object detection and semantic segmentation.

\section{Acknowledgements}

The authors express their sincere gratitude to Gintare Karolina Dziugaite (Google Brain) and Himanshu Arora (Workday) for providing valuable initial feedback in refining the idea, and to Ajay Arasanipalai (UIUC) for helping with code review and ablation experiments. 

\clearpage
%
%
\bibliographystyle{plainnat}
\bibliography{egbib}
\end{document}